\documentclass[runningheads]{llncs}
\usepackage[T1]{fontenc}
\usepackage{graphicx}
\usepackage{accvabbrv}

\usepackage{booktabs}
\usepackage{multirow} 
\usepackage{adjustbox}
\usepackage{tabularx}
\usepackage{caption, subcaption}

\usepackage{colortbl} 
\usepackage{xcolor}
\usepackage{stmaryrd}

\usepackage[accsupp]{axessibility}  % Improves PDF readability for those with disabilities.

\usepackage{orcidlink} 
\usepackage{hyperref}
\usepackage{color}

\begin{document}
\title{O1O: Grouping of Known Classes to Identify Unknown Objects as Odd-One-Out} 
%
% If the paper title is too long for the running head, you can set
% an abbreviated paper title here
\titlerunning{O1O}
%

% TODO FINAL: Replace with your author list. 
% Include the authors' OCRID for the camera-ready version, if at all possible.
\author{Mısra Yavuz\inst{}\orcidlink{0009-0004-9517-2506} \and
Fatma Güney\inst{}\orcidlink{0000-0002-0358-983X} 
% \and Third Author\inst{3}\orcidlink{2222--3333-4444-5555}
}
\authorrunning{M. Yavuz et al.}
% First names are abbreviated in the running head.
% If there are more than two authors, 'et al.' is used.
%
\institute{
Department of Computer Engineering, Koç University \\
KUIS AI Center\\
\email{\{myavuz21,fguney\}@ku.edu.tr}
}

\maketitle              % typeset the header of the contribution

\newcommand{\Perp}{\perp\!\!\! \perp}
\newcommand{\bK}{\mathbf{K}}
\newcommand{\bX}{\mathbf{X}}
\newcommand{\bY}{\mathbf{Y}}
\newcommand{\bk}{\mathbf{k}}
\newcommand{\bx}{\mathbf{x}}
\newcommand{\by}{\mathbf{y}}
\newcommand{\bhy}{\hat{\mathbf{y}}}
\newcommand{\bty}{\tilde{\mathbf{y}}}
\newcommand{\bG}{\mathbf{G}}
\newcommand{\bI}{\mathbf{I}}
\newcommand{\bg}{\mathbf{g}}
\newcommand{\bS}{\mathbf{S}}
\newcommand{\bs}{\mathbf{s}}
\newcommand{\bM}{\mathbf{M}}
\newcommand{\bw}{\mathbf{w}}
\newcommand{\eye}{\mathbf{I}}
\newcommand{\bU}{\mathbf{U}}
\newcommand{\bV}{\mathbf{V}}
\newcommand{\bW}{\mathbf{W}}
\newcommand{\bn}{\mathbf{n}}
\newcommand{\bv}{\mathbf{v}}
\newcommand{\bwv}{\mathbf{wv}}
\newcommand{\bq}{\mathbf{q}}
\newcommand{\bR}{\mathbf{R}}
\newcommand{\bi}{\mathbf{i}}
\newcommand{\bj}{\mathbf{j}}
\newcommand{\bp}{\mathbf{p}}
\newcommand{\bt}{\mathbf{t}}
\newcommand{\bJ}{\mathbf{J}}
\newcommand{\bu}{\mathbf{u}}
\newcommand{\bB}{\mathbf{B}}
\newcommand{\bD}{\mathbf{D}}
\newcommand{\bz}{\mathbf{z}}
\newcommand{\bP}{\mathbf{P}}
\newcommand{\bC}{\mathbf{C}}
\newcommand{\bA}{\mathbf{A}}
\newcommand{\bZ}{\mathbf{Z}}
\newcommand{\bff}{\mathbf{f}}
\newcommand{\bF}{\mathbf{F}}
\newcommand{\bo}{\mathbf{o}}
\newcommand{\bO}{\mathbf{O}}
\newcommand{\bc}{\mathbf{c}}
\newcommand{\bm}{\mathbf{m}}
\newcommand{\bT}{\mathbf{T}}
\newcommand{\bQ}{\mathbf{Q}}
\newcommand{\bL}{\mathbf{L}}
\newcommand{\bl}{\mathbf{l}}
\newcommand{\ba}{\mathbf{a}}
\newcommand{\bE}{\mathbf{E}}
\newcommand{\bH}{\mathbf{H}}
\newcommand{\bd}{\mathbf{d}}
\newcommand{\br}{\mathbf{r}}
\newcommand{\be}{\mathbf{e}}
\newcommand{\bb}{\mathbf{b}}
\newcommand{\bh}{\mathbf{h}}
\newcommand{\bhh}{\hat{\mathbf{h}}}
\newcommand{\btheta}{\boldsymbol{\theta}}
\newcommand{\bTheta}{\boldsymbol{\Theta}}
\newcommand{\bpi}{\boldsymbol{\pi}}
\newcommand{\bphi}{\boldsymbol{\phi}}
\newcommand{\bpsi}{\boldsymbol{\psi}}
\newcommand{\bPhi}{\boldsymbol{\Phi}}
\newcommand{\bmu}{\boldsymbol{\mu}}
\newcommand{\bsigma}{\boldsymbol{\sigma}}
\newcommand{\bSigma}{\boldsymbol{\Sigma}}
\newcommand{\bGamma}{\boldsymbol{\Gamma}}
\newcommand{\bbeta}{\boldsymbol{\beta}}
\newcommand{\bomega}{\boldsymbol{\omega}}
\newcommand{\blambda}{\boldsymbol{\lambda}}
\newcommand{\bLambda}{\boldsymbol{\Lambda}}
\newcommand{\bkappa}{\boldsymbol{\kappa}}
\newcommand{\btau}{\boldsymbol{\tau}}
\newcommand{\balpha}{\boldsymbol{\alpha}}
\newcommand{\nR}{\mathbb{R}}
\newcommand{\nN}{\mathbb{N}}
\newcommand{\nL}{\mathbb{L}}
\newcommand{\nE}{\mathbb{E}}
\newcommand{\cN}{\mathcal{N}}
\newcommand{\cM}{\mathcal{M}}
\newcommand{\cR}{\mathcal{R}}
\newcommand{\cB}{\mathcal{B}}
\newcommand{\cL}{\mathcal{L}}
\newcommand{\cH}{\mathcal{H}}
\newcommand{\cS}{\mathcal{S}}
\newcommand{\cT}{\mathcal{T}}
\newcommand{\cO}{\mathcal{O}}
\newcommand{\cC}{\mathcal{C}}
\newcommand{\cP}{\mathcal{P}}
\newcommand{\cE}{\mathcal{E}}
\newcommand{\cI}{\mathcal{I}}
\newcommand{\cF}{\mathcal{F}}
\newcommand{\cK}{\mathcal{K}}
\newcommand{\cY}{\mathcal{Y}}
\newcommand{\cX}{\mathcal{X}}
\def\bgamma{\boldsymbol\gamma}

\newcommand{\specialcell}[2][c]{%
  \begin{tabular}[#1]{@{}c@{}}#2\end{tabular}}

\newcommand{\figref}[1]{\Fig~\ref{#1}}
\newcommand{\secref}[1]{Section~\ref{#1}}
\newcommand{\algref}[1]{Algorithm~\ref{#1}}
\newcommand{\eqnref}[1]{Eq.~\eqref{#1}}
\newcommand{\tabref}[1]{Table~\ref{#1}}

\newcommand{\rulesep}{\unskip\ \vrule\ }

%\DeclareMathOperator*{\argmax}{argmax~}
% \DeclareMathOperator*{\argmin}{argmin~}

% KL divergence
%\DeclarePairedDelimiterX{\infdivx}[2]{[}{]}{%
%  #1\;\delimsize\|\;#2%
%}
%\newcommand{\infdiv}{D\infdivx}

% Kullback-Leibler divergence (or relative entropy)
\newcommand{\KLD}[2]{D_{\mathrm{KL}} \Big(#1 \mid\mid #2 \Big)}

\renewcommand{\b}{\ensuremath{\mathbf}}

\def\mc{\mathcal}
\def\mb{\mathbf}

\newcommand{\T}{^{\raisemath{-1pt}{\mathsf{T}}}}

\makeatletter
\DeclareRobustCommand\onedot{\futurelet\@let@token\@onedot}
\def\@onedot{\ifx\@let@token.\else.\null\fi\xspace}
\def\eg{e.g\onedot} \def\Eg{E.g\onedot}
\def\ie{i.e\onedot} \def\Ie{I.e\onedot}
\def\cf{cf\onedot} \def\Cf{Cf\onedot}
\def\etc{etc\onedot} \def\vs{vs\onedot}
\def\wrt{wrt\onedot}
\def\dof{d.o.f\onedot}
\def\etal{et~al\onedot} \def\iid{i.i.d\onedot}
\def\Fig{Fig\onedot} \def\Eqn{Eqn\onedot} \def\Sec{Sec\onedot} \def\Alg{Alg\onedot}
\makeatother

\newcommand{\xdownarrow}[1]{%
  {\left\downarrow\vbox to #1{}\right.\kern-\nulldelimiterspace}
}

\newcommand{\xuparrow}[1]{%
  {\left\uparrow\vbox to #1{}\right.\kern-\nulldelimiterspace}
}

% nice url font and color
% \renewcommand\UrlFont{\color{blue}\rmfamily}

% rotation
\newcommand*\rot{\rotatebox{90}}
\newcommand{\boldparagraph}[1]{\vspace{0.2cm}\noindent{\bf #1:} }
\newcommand{\boldquestion}[1]{\vspace{0.2cm}\noindent{\bf #1?} }

% colored tables
\definecolor{First}{HTML}{BDE6CD}%{8EE574}
\definecolor{Second}{HTML}{E2EEBC}%{FCFF91}
\definecolor{Third}{HTML}{FFF8C5}%{FFC5BF}
\definecolor{DarkBlue}{HTML}{89AFE2} % For blue
\definecolor{Green}{HTML}{66AB9F} % For green
\definecolor{Red}{HTML}{EB7470} % For red

%definecolor{First}{rgb}{1, 0.6, 0.6}
%\definecolor{Second}{rgb}{1, 0.8, 0.6}
%\definecolor{Third}{rgb}{1,1, 0.6}

% table utilities
\newcommand{\fst}[1]{\cellcolor{First}#1}
\newcommand{\snd}[1]{\cellcolor{Second}#1}
\newcommand{\trd}[1]{\cellcolor{Third}#1}

\newcommand{\my}[1]{ \noindent {\color{blue} {\bf Misra:} {#1}} } 
\newcommand{\nn}[1]{ \noindent {\color{green} {\bf nazir:} {#1}} }
\newcommand{\ftm}[1]{ \noindent {\color{cyan} {\bf Fatma:} {#1}} }
\begin{abstract}
Object detection methods trained on a fixed set of known classes struggle to detect objects of unknown classes in the open-world setting. 
Current fixes involve adding approximate supervision with pseudo-labels corresponding to candidate locations of objects, typically obtained in a class-agnostic manner. 
While previous approaches mainly rely on the appearance of objects, we find that geometric cues improve unknown recall. 
Although additional supervision from pseudo-labels helps to detect unknown objects, it also introduces confusion for known classes. 
We observed a notable decline in the model's performance for detecting known objects in the presence of noisy pseudo-labels. 
Drawing inspiration from studies on human cognition, we propose to group known classes into superclasses. 
% By doing so, our model can identify similarities between classes within a superclass, aiding in detecting unknown classes through an odd-one-out scoring mechanism. 
By identifying similarities between classes within a superclass, we can identify unknown classes through an odd-one-out scoring mechanism. Our experiments on open-world detection benchmarks demonstrate significant improvements in unknown recall, consistently across all tasks. 
Crucially, we achieve this without compromising known performance, thanks to better partitioning of the feature space with superclasses.
Project page: \url{https://kuis-ai.github.io/O1O}.


\end{abstract}
\section{Introduction}
\label{sec:intro}

Object detection methods in open-world settings are required to recognize objects from both known and unknown classes~\cite{Bendale2015CVPR,Joseph2021CVPR}. 
This task is particularly challenging due to having only a static dataset with a limited set of classes to learn from.
Without any effort to recognize unknown classes, the standard object detection approach partitions the representation space between the known classes. Since unseen classes are not represented in the learned space, this approach generalizes poorly to the open-world setting. To localize objects of unseen classes, open-world detectors typically resort to additional information.

\begin{figure}[t]
  \centering
  \includegraphics[width=\linewidth]{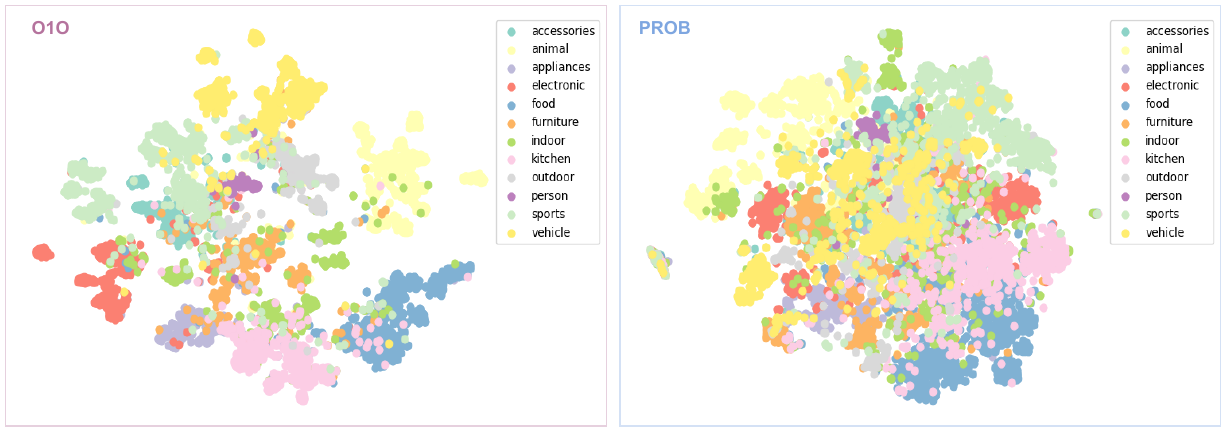}
  \caption{\textbf{Superclass Prior \vs Single Objectness Distribution.} 
  This figure compares the projections of object queries trained with superclass prior in our approach O1O (\textbf{left}) \vs fitting a single objectness distribution to matched queries as in PROB~\cite{Zohar2023CVPR} (\textbf{right}) for S-OWOD Task 4. O1O shapes the representation space by encouraging queries of similar classes to group together, allowing us to identify odd-one-out queries as unknown objects.
  }
  \label{fig:feat_space}
\end{figure}

In object detection, both the localization and classification are supervised using ground truth annotations for known classes. In the case of unknown classes, however, there is no ground truth data available for training, mimicking the real-world setting. The lack of data has led to the mining of unknown objects from highly activated areas in the background~\cite{Gupta2022CVPR} or by using proposals of object localization networks~\cite{Fang2023ARXIV}. 
This pseudo-labeling strategy generates approximate ground truth for unknown objects, enabling the model to search for objects beyond known classes.
However, pseudo-labels may not necessarily correspond to object regions, and directly using noisy targets without any safeguarding degrade the performance of known classes as we show in our experiments.

A related line of work called class-agnostic object detection studies object localization without classification, \ie separating foreground from background \cite{Maaz2021ECCV,Zhao2022ECCV,Kim2021RAL,Saito2022ECCV,Maaz2021ARXIV,Huang2023ICLR}. 
Recent work~\cite{Huang2023ICLR} 
% reports 
improves recall by incorporating geometric cues such as predicted depth and surface normal maps into region proposal networks (RPN)~\cite{Ren2015PAMI}. However, the potential of geometric cues remains unexplored in open-world object detection (OWOD) where the goal is not only to improve unknown performance but also to maintain known performance while learning novel categories incrementally. 
As a first step, 
% in that direction, 
we trained a state-of-the-art OWOD model, PROB~\cite{Zohar2023CVPR}, with geometric pseudo-labels. PROB~\cite{Zohar2023CVPR} learns a class-agnostic distribution to represent generic objectness. 
This essentially pushes all matched object queries towards an objectness center, forcing them to be similar to each other~(\figref{fig:feat_space}). This approach degrades known performance due to noisy pseudo-labels that may not correspond to an object.

Learning a single distribution to represent any object is too simple to capture the diverse characteristics of object groups and is expected to perform poorly as the space of known concepts gets larger (\figref{fig:feat_space}). The other extreme is learning class prototypes by fitting a distribution to each known class~\cite{Du2022NeurIPS}, which is challenging because of the varying frequencies and specifics of each class, and creates more confusion for the model. To learn better objectness priors, we draw inspiration from human cognitive science. Studies in human perception suggest that humans use common attributes such as material, shape, color, and functional properties to recognize and describe objects~\cite{Hebart2020revealing}.
Inspired by this idea, we propose a middle ground between these two extremes: grouping similar object classes into superclasses. We learn a discriminative classifier for each known superclass, such as \textit{person}, \textit{vehicle}, or \textit{furniture}. 
By partitioning the query embedding space into superclasses, we recalibrate the confidence scores for known classes with a conditional probability formulation. Our experiments demonstrate that recalibrated classification scores prevent degradation in the performance of known classes when using pseudo-labels for unknown classes, and achieve the best balance between known and unknown performance.

\section{Related Work}
\label{sec:rw}

\subsection{Open-World Object Detection} 
The problem of detecting unknown objects has been tackled early on as an open-set recognition problem where the goal is to identify concepts that are semantically disjoint from the training set labels~\cite{Miller2017ICRA,Hall2018WACV,Miller2018ICRA,Dhamija2020WACV,Miller2021RAL,Han2022CVPR,Zheng2022CVPRW}. Simultaneously, methods pioneered by Joseph \etal~\cite{Joseph2021CVPR} extended this formulation to open-world object detection (OWOD). In OWOD, the aim is to detect unknown objects and incrementally learn novel categories without compromising the performance of previously known classes. This setting aligns more closely with real-life requirements in ever-changing environments. Existing OWOD methods follow different approaches to separate objects from the background, whether known or unknown. Some methods extract pseudo-labels, \ie candidate bounding boxes for unknown objects as auxiliary supervision~\cite{Joseph2021CVPR,Wu2022ECCV}, while others directly predict an objectness score~\cite{Gupta2022CVPR,Zohar2023CVPR,Ma2023CVPR}. 
Generally, OWOD methods can be categorized based on their pseudo-labeling approach and modeling of knowns. 

\boldparagraph{RPN-based methods} 
Earlier approaches build on two-stage architectures in the R-CNN family~\cite{Girshick2013CVPR,Girshick2015ICCV,Ren2015PAMI}. 
Commonly, these methods~\cite{Joseph2021CVPR,Wu2022ECCV} utilize the region proposal network (RPN) proposals to model unknowns. 
% Once the proposals are generated, 
The classification head is extended to account for unknown objects.
Building on~\cite{Joseph2021CVPR}, Yu \etal~\cite{Yu2022ICIP} learn class-specific prototypes. 
Wu \etal~\cite{Wu2022HCMA@MM} attach a branch to learn objectness by predicting the IoU score of a proposal as an indicator of localization quality. In contrast to RPN proposals, unsupervised methods can be used to alleviate the bias of known classes on pseudo-labels.
Wang \etal~\cite{Wang2023ICCV} first randomly generate bounding box proposals and then propose a matching score to select the ones that are most likely foreground unknown objects. In an autoencoding paradigm, Fang \etal~\cite{Fang2023ARXIV} learn separate distributions for foreground and background to identify objects better, and use selective search~\cite{Uijlings2013IJCV} for pseudo-labels.

\boldparagraph{DETR-based methods} More recent work builds on the transformer-based architecture DETR~\cite{Zhu2021ICLR}. 
Absent an RPN module, these methods explore various pseudo-label sources. Gupta \etal~\cite{Gupta2022CVPR} aim to capture unknowns with a scoring function based on the attention activation of intermediate layers. They use two classification heads: one to separate knowns from unknowns and another to distinguish foreground from background. Ma \etal~\cite{Ma2023CVPR} combine selective search~\cite{Uijlings2013IJCV} with attention activations~\cite{Gupta2022CVPR} to generate pseudo-labels. They decouple localization and classification in a cascade fashion to minimize the known class bias.

Zohar et al.~\cite{Zohar2023CVPR} take a different approach by omitting pseudo-labels and instead learning a distribution of objectness through joint optimization of parameters for a multivariate Gaussian on matched object queries from DETR. He \etal~\cite{He2023ARXIV} uses a similar modeling of objectness but based on the early decoder layers to mitigate the influence of class-specific features. Additionally, they generate pseudo-labels using the Segment Anything Model (SAM)~\cite{Kirillov2023ICCV}, which significantly contributes to their performance on unknown objects.
While it is intriguing to utilize foundational models for OWOD, the definition of unknown becomes blurry for SAM, which has been trained on a large collection of internet images. 
Our method also builds on the variations of DETR. Differently, we explore the potential of geometric cues to detect unknowns.

\subsection{Class-Agnostic Object Detection} 
In class-agnostic object detection, some methods utilize the rich semantics of natural language to ease localization~\cite{Maaz2021ECCV,Zhao2022ECCV}. Saito \etal~\cite{Saito2022ECCV} augment images by synthetically pasting foreground objects to alleviate the bias from unlabelled foreground objects in existing datasets. OLN~\cite{Kim2021RAL} estimates the objectness of proposals based on localization quality supervised by ground truth. GOOD~\cite{Huang2023ICLR} argues that geometric cues carry more relevant information about objectness that generalizes across categories. We explore the potential of GOOD~\cite{Huang2023ICLR} as a source of structured objectness priors for open-world object detection.

\subsection{Superclass Supervision}
The idea of a semantic grouping of classes has previously been used in the standard object detection task to improve closed-set performance. 
Specifically, SGNet~\cite{li2021sgnet} trains a superclass classification branch in an end-to-end fashion with a convolutional detection model. 
Additionally, long-tail classification methods~\cite{zhou2018deep,du2023superdisco} explored the superclass grouping to enhance recognition of tail classes by benefiting from less imbalance in superclass distributions.
Differently, we encourage objects sharing similar properties to cluster within a superclass by classifying queries into superclasses. 
Through re-calibration of confidence scores and a novel unknown scoring function, we identify unknown objects that do not share these properties as odd-one-out (O1O). We are the first to explore the potential of superclasses for unknown object detection.
\section{Methodology}
\label{sec:method}
\begin{figure}[t]
  \centering
  \includegraphics[width=\linewidth]{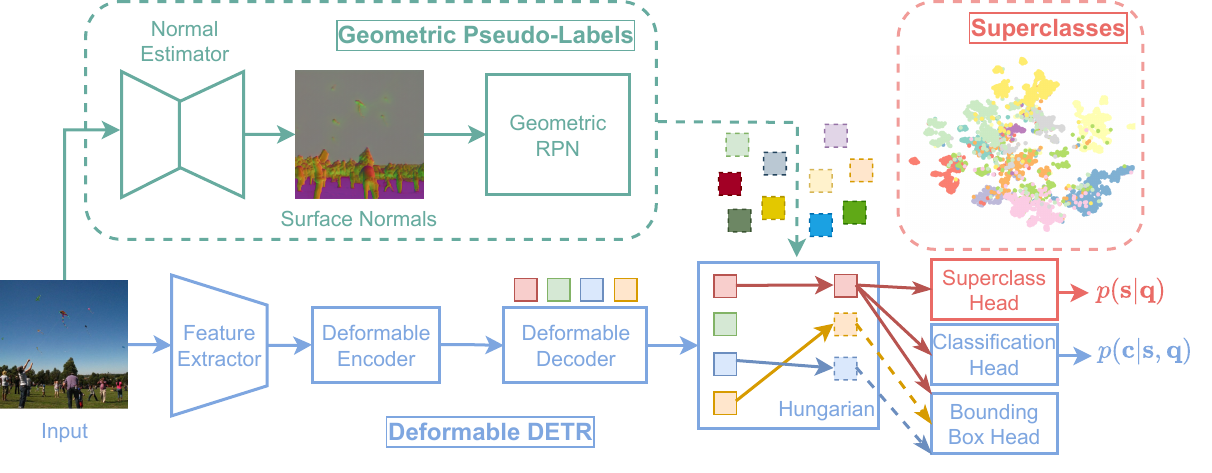}
    \caption{{\textbf{Overview of Proposed O1O.} Building on Deformable DETR~\cite{Zhu2021ICLR} \textcolor{DarkBlue}{(blue)}, we first add supervision for unknowns with geometric pseudo-labels. Following GOOD~\cite{Huang2023ICLR}, we extract pseudo-labels \textcolor{black}{(dashed)} from a Region Proposal Network (RPN) trained on surface normal maps \textcolor{Green}{(green)}. This allows us to localize unknown objects based on geometric cues in a class-agnostic manner. Noisy pseudo-labels tend to hurt the known performance of a model. 
    To mitigate this issue, we propose to group queries into superclasses with a superclass head \textcolor{Red}{(red)}. By incorporating the learned prior from superclasses into the scoring function, we achieve the best balance between known and unknown performance.}}
  \label{fig:pipeline}
\end{figure}

We introduce Odd-One-Out (O1O), an open-world object detection method to locate and classify objects by identifying unknowns, as illustrated in \figref{fig:pipeline}. 
We first provide background on the transformer-based architectures for object detection including the extensions that we adopt in \secref{sec:detr}. Next, we introduce geometric cues to obtain pseudo-labels for unknowns in \secref{sec:good}.
For known performance, we propose to group known classes into superclasses and formulate classification as a conditional probability distribution in \secref{sec:superclass}. Lastly, in \secref{sec:inc_learn}, we explain our incremental learning strategy for open-world tasks.

\subsection{Background on DETR}
\label{sec:detr}
DEtection TRansformer (DETR)~\cite{Carion2020ECCV} operates by reasoning about the relationships between objects and their interaction with image features using a transformer as shown in blue in \figref{fig:pipeline}. Formally, it initializes $N_{query}$ learned positional encoding vectors. In the transformer encoder, image features interact with each other and produce weighted attention maps. In the transformer decoder, queries first interact with each other and then with the feature maps from the encoder. This results in an embedding $\bq_i \in \nR^{d}$ for each object query $i \in N_{query}$, incorporating information about other queries and the image context. Each object query is then independently decoded into box coordinates and class labels using a feedforward network (FFN) followed by a linear layer for output generation. During training, ground truth objects are matched to object queries via bipartite matching before computing the classification and bounding box losses. Queries unmatched with ground truth objects are responsible for predicting a special class label, typically denoted as K+1, which in closed-set methods represents the `no-object' case, while in open-set methods, it is used for the unknown class. 
Note that unmatched queries do not receive any supervision for localization.

\boldparagraph{Extensions} Deformable DETR~\cite{Zhu2021ICLR} extends DETR by incorporating a more efficient attention module known as deformable attention. This attention mechanism focuses on a small set of key sampling points around a 2D reference point, enhancing the model's ability to attend to relevant image features. Leveraging this efficiency, Deformable DETR integrates multi-scale image features, particularly beneficial for detecting small objects. Deformable DETR significantly improves the convergence capabilities of DETR. Another variant DAB-DETR~\cite{Liu2021ICLR} improves the cross-attention operation by employing dynamically updated 4D anchor boxes consisting of a reference point $(x, y)$ and a reference anchor size $(w, h)$. The more recent DN-DETR~\cite{Li2022CVPR} further improves the convergence with denoising techniques during training. DN-DETR identifies bipartite matching as the bottleneck for slow convergence and mitigates it by injecting noisy ground truth boxes into the transformer decoder and then training the model to reconstruct the original boxes. This approach effectively halves the number of training epochs required to match the performance of Deformable DETR.

While most recent OWOD methods~\cite{Gupta2022CVPR,Zohar2023CVPR} build on Deformable DETR~\cite{Zhu2021ICLR}, we incorporate improvements from both DN-DETR~\cite{Li2022CVPR} and DAB-DETR~\cite{Liu2021ICLR}. Our primary objective is to leverage enhanced convergence properties for rapid experimentation while maintaining the overall performance.
Note that these architectural modifications improve efficiency but do not contribute to unknown performance. For fairness, we report the performance of the SOTA unknown object detection method, PROB~\cite{Zohar2023CVPR} using the same architecture (See \secref{sec:ablation}.)

\subsection{Pseudo-Labels for Unknown Objects}
\label{sec:good}
In DETR-based methods for unknown object detection, the class index $K+1$ represents the unknown class in addition to the $K$ known classes. 
Since unmatched queries only receive semantic supervision, the model's ability to locate objects beyond the known set is limited. The main challenge is obtaining labels for localizing unknown objects. Previous methods employ pseudo-labeling strategies to obtain an auxiliary signal for training to localize all objects. % instead of only known ones. 

\boldparagraph{Geometric Cues} Recent work in class-agnostic object localization, proposes to use geometric cues for open-world object detection~\cite{Huang2023ICLR}. While RGB-based methods suffer from over-fitting to semantic features of known classes, geometric cues can be used to mine novel object classes as pseudo-labels. Specifically, GOOD~\cite{Huang2023ICLR} trains a region proposal network (RPN) using geometric cues such as depth and surface normal maps as input. By focusing on object shapes, relative spatial locations (depth), and directional changes (normal vectors), GOOD can retrieve more accurate pseudo-labels compared to other class-agnostic detectors. 

In this work, we use pseudo-labels obtained from GOOD's RPN module to train for unknown object detection. 
For fairness, we adhere closely to OWOD benchmark settings regarding data partitioning into known and unknown classes. For training the RPN, we ensure that there are no objects of unknown classes according to the benchmark setting considered. This cannot be ensured while using sources like SAM~\cite{Kirillov2023ICCV} that is trained on a large collection of internet images. 

\subsection{Shaping Queries Through Superclass Supervision}
\label{sec:superclass}
Following the standard DETR-based approach, our model takes an image as input and outputs bounding box locations and class scores corresponding to objects on the image. 
Let $\bc \in \{0, 1\}^{K+1}$ be a random variable representing the one-hot encoded class of a bounding box over a predefined set of $K$ known classes and the unknown label $K+1$. Based on the embedding of query vectors $\{\bq_i\}_{i=1}^{N_{query}}$, we re-design the classification head.

We classify regions by further decomposing the probability of a class with superclasses. 
A superclass is a group of classes that share some semantic properties. 
% For example; cats, dogs, and birds can be grouped as \textit{animals}, and cars, buses, and motorcycles can be grouped as \textit{vehicles}.
For example; cats, dogs, and birds as \textit{animals}, and cars, buses, and motorcycles as \textit{vehicles}.
In our approach, we propose to learn common properties between known classes by grouping them into superclasses. 
That way, we can better judge similarities between known classes, \ie recalibrating the confidence of what is known, and then use it to accurately identify what is unknown.

\boldparagraph{Superclass Prior} 
% In DETR, the per-class classification head outputs the probability of belonging to a class $\bc$ for each query $\bq$ as $p(\bc \vert \bq)$.
In DETR, the per-class classification head outputs the probability of each query $\bq$ belonging to a class $\bc$ as $p(\bc \vert \bq)$. 
In addition to that, we train a superclass head to group known classes into superclasses.
Formally, the superclass head outputs a probability of belonging to a superclass $\bs$ as $p(\bs \vert \bq)$ over the set of superclasses $S$. For each known class, we recalibrate its confidence score by multiplying it with the probability of the superclass that it belongs to:

\begin{equation}
    p'(\bc \vert \bq) = \sum_{\bs \in S} p(\bc \vert \llbracket \bc \in \bs \rrbracket, \bq) \cdot p(\bs \vert \bq) 
\label{eq:score}
\end{equation}
where $\llbracket \cdot \rrbracket$ denotes Iverson bracket. Since known classes are fixed, we can pre-define the set of superclasses based on the known classes (see Supplementary). In practice, this reduces conditioning on the superclass $\bs$ in the first term to a lookup operation to find the corresponding superclass for a class. Intuitively, we shape the representation space of known classes by encouraging similar properties for queries matched to classes within a superclass.

\boldparagraph{Unknown Probability} For unknown probability $p(u \vert \bq)$, we apply the odd-one-out approach and define its probability as the complement event of being known. Considering each class probability to be independent, we sum the recalibrated probabilities of belonging to known classes, and subtract it from 1:
\begin{equation}
    p(u \vert \bq) = 1 - \sum_{\bc \in K} p'(\bc \vert \bq) 
\label{eq:unk_score}
\end{equation}
Essentially, the decoupled conditional probability helps the model to update its belief for a query about belonging to a class by considering the similarities with its superclass. If the model is not confident enough to label a query as known, then \eqref{eq:unk_score} becomes higher, and the query is predicted as unknown. In our experiments, we show that learning a structured prior with superclasses recovers the known performance by reducing the confusion caused by noisy supervision from pseudo-labels for unknown classes. 

\boldparagraph{Training} 
% We first train the RPN on known classes only on the image set of the current task and store the generated proposals. Then, we train our model end-to-end on a weighted combination of localization, per-class classification, and superclass losses: 
We first train the RPN on only the known set of the current task and store the generated proposals. Then, we train our model on a weighted combination of localization, per-class classification, and superclass losses: 

\begin{align}
    \cL =  \underbrace{\cL_{\text{bbox}} + \cL_{\text{giou}}}_\text{localization} 
    + \underbrace{\cL_{\text{cls}} + {\color{blue} \cL_{\text{sup}}}}_{\text{classification}}
\end{align}
We use the same loss function used in DETR-variants and only add the classification loss $\cL_{\text{sup}}$ (shown in blue) to train the superclass head. We use focal loss~\cite{Lin2017ICCV} for superclass classification to address the imbalance issue in the distribution of superclasses.
We apply the classification losses only on queries matched to known targets. In other words, we only apply the localization losses for the queries matched to pseudo-labels. 
For the classification of unknowns, we use the knowledge of the model on known classes to reason about what is unknown. 

Building on extended Deformable DETR with denoising~\cite{Li2022CVPR}, we apply a denoising loss for each component, including the superclass loss. For noisy bounding box queries, the denoising bounding box loss reconstructs original coordinates, while denoising classification losses predict the original target labels.

\boldparagraph{Inference} In inference, we use the decomposition in \eqref{eq:score} to obtain classification confidence scores of known classes. Since one class only belongs to a specific superclass, and this matching is given in the label set, it serves as a prior rather than privileged information. 
% We only use the currently known class label set to access superclass information, and unseen objects are considered to belong to an additional unknown superclass. 
We use only the current known class label set to access superclass information, with unseen objects considered part of an additional unknown superclass.
% For the unknown score, we use \eqref{eq:unk_score} and decide a threshold on a subset of the training set, such that the 95\% of known instances would be classified correctly.
For the unknown score, we use \eqref{eq:unk_score} and set a threshold on a training subset to ensure 95\% of known instances are classified correctly.

\subsection{Incremental Learning}
\label{sec:inc_learn}
Apart from unknown detection, OWOD formulation requires incrementally learning novel classes while still maintaining the performance on previously learned concepts. 
However, training a model continually on multiple tasks typically results in catastrophic forgetting~\cite{Mccloskey1989}. 
To prevent such a phenomenon, the common practice is to store a small set of exemplars 
% which is then used to fine-tune the model on previous tasks. 
for fine-tuning the model on previous tasks.
Most OWOD approaches select the exemplar set randomly. On the other hand, an active selection strategy can be applied for a more effective replay. 
For example, PROB~\cite{Zohar2023CVPR} uses their objectness approximation to actively select images with high and low-scoring instances, representing easy and hard cases. 

We adopt a similar active exemplar selection strategy using our class conditional probability \eqref{eq:score}. 
Specifically, we calculate the class-wise conditional probabilities for all queries on the training set of the previous task. Then, for each class, we select the images containing the highest 25 and the lowest 25 scoring instances and save them as exemplars to be used in the next tasks. 
If the number of selected images turns out to be higher than in previous work, we randomly drop out the extra number of images to be comparable.
\section{Experiments}
\label{sec:exp}
\subsection{Benchmarks} 
We report results on two OWOD benchmarks. M-OWOD, proposed by~\cite{Joseph2021CVPR}, is composed of both MS-COCO and PASCAL images. 80 categories of the MS-COCO dataset are split into four non-overlapping tasks ${\{T_1, \dots, T_4\}}$ with 20 classes being introduced at each. In each task $T_t$, only the annotations of classes that have been introduced until $T_t$ are used for training, and the rest are removed. The test set contains instances from all 80 classes, and the known \vs unknown performance is calculated according to the known-unknown split. S-OWOD, proposed by \cite{Gupta2022CVPR}, only consists of MS-COCO images. Classes are split according to the supercategories they belong to. For example, all animal classes are included in Task 1, while all sports classes are introduced in Task 3. Denoting the differences, these benchmarks are also known as Mixed-superclass and Separated-superclass, respectively. For more details, please refer to \cite{Joseph2021CVPR} and \cite{Gupta2022CVPR}. 

\subsection{Evaluation Metrics} 
Following the common practice in the OWOD literature, we report the mean average precision (mAP) for known performance. Since not all objects are annotated in any given dataset, unknown precision is ill-defined. A model might detect an object in a location where a real object exists, but if it does not belong to the predefined categories in the dataset, it might have been excluded from the ground truth annotations. Therefore, recall (U-Recall) is used as the main metric to evaluate the ability to retrieve unknown objects. 

\subsection{Implementation Details}  
We build our method on the extended version of Deformable DETR~\cite{Zhu2021ICLR} with denoising and 4D anchor boxes~\cite{Liu2021ICLR,Li2022CVPR}. We use the same backbone as the prior work, \ie ResNet-50 FPN~\cite{He2015CVPR} pre-trained on ImageNet~\cite{Deng2009CVPR} in a self-supervised manner~\cite{Caron2021ICCV}. We set the number of queries to 100 and the query embedding dimension to 256. We use predicted surface normal maps to train RPN after comparing different sources, including RGB images and predicted depth maps (see Supplementary). To select pseudo-boxes, we threshold RPN's confidence scores at 0.5 and apply batched NMS on ground truth and pseudo annotations with an IoU threshold of 0.5 to remove duplicates. 

We train our model on 4 GPUs, with a batch size of 6, and 2e-4 as the starting learning rate. Since Deformable DETR with the extensions converges fast, we drop the learning rate by 10 at epoch 18. We train for 30 epochs on M-OWOD Task 1 and 20 epochs on S-OWOD Task 1. Since the dataset of S-OWOD Task 1 is five times larger, our model converges earlier. 
For incremental tasks, we train for 10 epochs with a learning rate of 2e-5. We perform fine-tuning from exemplar replay for 20 epochs, with a learning rate of 2e-4 for the first half and 2e-5 for the second half. Overall, our architectural improvements cut down training time approximately by half compared to previous DETR-based approaches.

\subsection{Open-World Object Detection Performance}
\label{sec:results}

In \tabref{tab:owod_results}, we present O1O's performance in OWOD, comparing it to previous approaches across different tasks on two benchmarks. O1O consistently outperforms all methods across all tasks on both benchmarks except for USD~\cite{He2023ARXIV} on S-OWOD in terms of unknown performance (U-Recall), while maintaining a competitive or the highest Known AP. 
Note that there's a trade-off between known and unknown performance, as each method can make a fixed number of predictions in total. While increasing unknown recall by predicting more unknowns is possible, it would inevitably degrade known performance. Therefore, striking a balance between these two metrics is crucial. 

\boldparagraph{S-OWOD \vs M-OWOD} While the two benchmarks differ in their partitioning of superclasses, our method demonstrates the ability to generalize to both, as indicated by the results. Due to the superclass separation between tasks in S-OWOD, learning becomes easier, resulting in better-known performance for all methods compared to M-OWOD. The larger dataset size of Task 1 on S-OWOD also contributes to higher known detection performances on this benchmark.

\begin{table*}[t!]
    \centering
    \adjustbox{max width=\textwidth}{%
        \begin{tabular}{l | cc | cccc | cccc | ccc}
            \toprule
            Task IDs & \multicolumn{2}{c|}{Task 1} & \multicolumn{4}{c|}{Task 2} & \multicolumn{4}{c|}{Task 3} & \multicolumn{3}{c}{Task 4} \\
            % \cmidrule(lr){2-3} \cmidrule(lr){4-7} \cmidrule(lr){8-11} \cmidrule(lr){12-14}
            \toprule
            \multirow{3}{*}{Methods} & U-Recall  & mAP $(\uparrow)$ & U-Recall & \multicolumn{3}{c|}{mAP $(\uparrow)$} & U-Recall & \multicolumn{3}{c|}{mAP $(\uparrow)$} & \multicolumn{3}{c}{mAP $(\uparrow)$} \\
            & \multirow{2}{*}{$(\uparrow)$} & Current  & \multirow{2}{*}{$(\uparrow)$} & Previously & Current &  \multirow{2}{*}{Both}   &  \multirow{2}{*}{$(\uparrow)$}  & Previously & Current & \multirow{2}{*}{Both}  & Previously & Current & \multirow{2}{*}{Both}  \\
            & & known  & & known & known &   &  & known & known &  & known & known &  \\
            \midrule
            ORE \textcolor{gray}{- ebui}~\cite{Joseph2021CVPR}  & 4.9 & 56.0 & 2.9 & 52.7 & 26.0 & 39.4 & 3.9 & 38.2 & 12.7 & 29.7 & 29.6 & 12.4 & 25.3 \\
            UC-OWOD~\cite{Wu2022ECCV}  & 2.4 & 50.7 & 3.4 & 33.1 & 30.5 & 31.8 & 8.7 & 28.8 & 16.3 & 24.6 & 25.6 & 15.9 & 23.2 \\
            OCPL~\cite{Yu2022ICIP}  & 8.26 & 56.6 & 7.65 & 50.6 & 27.5 & 39.1 & 11.9 & 38.7 & 14.7 & 30.7 & 30.7 & 14.4 & 26.7 \\
            2B-OCD~\cite{Wu2022HCMA@MM}  & 12.1 & 56.4 & 9.4 & 51.6 & 25.3 & 38.5 & 11.6 & 37.2 & 13.2 & 29.2 & 30.0 & 13.3 & 25.8 \\
            OW-DETR~\cite{Gupta2022CVPR}  & 7.5 & 59.2 & 6.2 & 53.6 & \trd{33.5} & 42.9 & 5.7 & 38.3 & 15.8 & 30.8 & 31.4 & 17.1 & 27.8 \\
            % OW-DETR (re-eval) ~\cite{Gupta2022CVPR}  & 6.0 & 58.5 & 6.7 & 52.8 & 6.2 & 29.5 & 7.5 & 27.0 & 15.8 & 23.3 & 1.1 & 15.4 & 4.7 \\
            % OW-DETR (nodup)~\cite{Gupta2022CVPR}  & 7.8 & 65.2 & 9.2 & 58.6 & 10.4 & 34.5 & 9.5 & 31.4 & 22.2 & 28.3 & 1.1 & 21.2 & 6.2 \\
            RandBox~\cite{Wang2023ICCV} & 10.6 & \trd{61.8} & 6.3 & - & - & \trd{45.3} & 7.8 & - & - & \trd{39.4} & - & - & \trd{35.4} \\

            PROB~\cite{Zohar2023CVPR} & 19.4 & 59.5 & 17.4 & 55.7 & 32.2 & 44.0 & 19.6 & \trd{43.0} & \trd{22.2} & 36.0 & \trd{35.7} & \trd{18.9} & 31.5 \\

            % CAT \textcolor{gray}{- cddw}& 19.1 & 59.3 & 18.6 & 52.8 & 30.2 & 41.5 & 21.0 & 41.0 & 17.6 & 33.0 & 32.6 & 15.8 & 27.9 \\
            CAT~\cite{Ma2023CVPR} & 23.7 & 60.0 & 19.1 & 55.5 & 32.7 & 44.1 & 24.4 & \trd{42.8} & 18.7 & 34.8 & 34.4 & 16.6 & 29.9 \\
            MEPU-FS~\cite{Fang2023ARXIV} & \trd{31.6} & 60.2 & \trd{30.9} & \trd{57.3} & \trd{33.3} & \trd{44.8} & \trd{30.1} & \trd{42.6} & 21.0 & 35.4 & 34.8 & \trd{19.1} & 30.9 \\
            MEPU-SS~\cite{Fang2023ARXIV} & 30.3 & 60.0 & \trd{30.6} & \trd{57.0} & \trd{33.1} & 44.5 & \trd{30.0} & 42.2 & 20.5 & 35.0 & 34.3 & \trd{18.9} & 30.4 \\
            % OW-RCNN & 37.7 & 63.0 & 39.9 & 48.8 & 41.7 & 45.2 & 43.0 & 45.2 & 31.7 & 40.7 & 40.3 & 28.8 & 37.4 \\
            USD \textcolor{gray}{- asf}~\cite{He2023ARXIV} & 21.6 & 59.9 & 19.7 & 56.6 & 32.5 & 44.6 & 23.5 & \snd{43.5} & \trd{21.9} & 36.3 & \trd{35.4} & \trd{18.9} & 31.3 \\
            USD~\cite{He2023ARXIV} & \snd{36.1} & 58.4 & \snd{34.7} & 54.3 & 31.4 & 42.7 & \snd{33.3} & 41.5 & 20.5 & 34.5 & 33.4 & 16.6 & 29.2 \\
            PROB${^\dagger}$~\cite{Zohar2023CVPR} & 28.3 & \fst{66.4} & 26.4 & \fst{62.6} & \snd{39.2}  & \snd{50.9}& 29.3 & \fst{49.6} & \snd{33.5} & \snd{44.2} & \snd{44.0} & \snd{26.5} & \snd{39.7} \\
            \midrule 
            % O1O (Ours) & 33.4 & 58.4 & 35.0 & 52.6 & 35.5 & 44.0 & 34.5 & 42.3 & 23.4 & 36.0 &  36.2  & 20.0 & 32.1 \\
            % OW-DETR${^\dagger}$~\cite{Gupta2022CVPR} & 10.1 & 65.6 & & & & & & & & & & & \\
            O1O${^\dagger}$ (Ours) & \fst{49.3} & \snd{65.1} & \fst{50.3} & \snd{61.0} & \fst{45.0} & \fst{53.0} & \fst{49.5} & \fst{50.0} & \fst{36.5} & \fst{45.5} & \fst{46.2} & \fst{31.0} & \fst{42.4} \\
            
            \midrule \midrule
            ORE \textcolor{gray}{- ebui}~\cite{Joseph2021CVPR}  & 1.5 & 61.4 & 3.9 & 56.5 & 26.1 & 40.6 & 3.6 & 38.7 & 23.7 & 33.7 & 33.6 & 26.3 & 31.8 \\
            OW-DETR~\cite{Gupta2022CVPR}  & 5.7 & 71.5 & 6.2 & 62.8 & 27.5 & 43.8 & 6.9 & 45.2 & 24.9 & 38.5 & 38.2 & 28.1 & 33.1 \\
            PROB~\cite{Zohar2023CVPR} & 17.6 & \snd{73.4} & 22.3 & \snd{66.3} & 36.0 & 50.4 & 24.8 & 47.8 & 30.4 & 42.0 & 42.6 & 31.7 & 39.9 \\
            CAT~\cite{Ma2023CVPR} & 24.0 & \fst{74.2} & 23.0 & \fst{67.6} & 35.5 & \trd{50.7} & 24.6 & \fst{51.2} & 32.6 & \trd{45.0} & \snd{45.4} & \snd{35.1} & \snd{42.8} \\
            MEPU-FS~\cite{Fang2023ARXIV} & \trd{37.9} & \fst{74.3} & \trd{35.8} & \fst{68.0} & \snd{41.9} & \fst{54.3} & \trd{35.7} & \snd{50.2} & \snd{38.3} & \snd{46.2} & \trd{43.7} & \trd{33.7} & \trd{41.2} \\
            MEPU-SS~\cite{Fang2023ARXIV} & 33.3 & \fst{74.2} & 34.2 & \fst{67.5} & \trd{41.0} & \snd{53.6} & 33.6 & \snd{50.0} & \trd{37.5} & \snd{45.8} & \trd{43.2} & \trd{33.5} & \trd{40.8} \\
            USD \textcolor{gray}{- asf}~\cite{He2023ARXIV} & 19.2 & \trd{72.9} & 22.4 & \trd{64.9} & 38.9 & \trd{51.2} & 25.4 & \snd{50.1} & 34.7 & \trd{45.0} & \trd{43.4} & \trd{33.6} & \trd{40.9} \\
            USD~\cite{He2023ARXIV} & \fst{51.1} & 69.8 & \fst{52.1} & 60.9 & 33.8 & 46.7 & \fst{49.9} & 45.8 & 30.2 & 40.6 & 39.0 & 28.7 & 36.4 \\
            \midrule 
            % O1O (Ours) & \fst{53.2} & \snd{73.5} & \fst{54.1} & \trd{64.7} & \fst{43.6} & \snd{53.6} & \fst{54.6} & \trd{48.3} & \fst{39.4} & \snd{45.3} & \fst{47.3} & \fst{42.0} & \fst{45.9} \\
            O1O (Ours) & \snd{49.8} & \trd{72.6} & \snd{51.1} & \trd{65.3} & \fst{44.9} & \fst{54.6} & \snd{48.1} & \trd{49.5} & \fst{41.5} & \fst{46.8} &  \fst{47.3} & \fst{42.0} & \fst{45.9} \\
            \bottomrule 
        \end{tabular}
    }
    \caption{\textbf{Comparison on the OWOD Benchmarks.} We compare our method O1O to the state-of-the-art on M-OWOD (\textbf{top}) and S-OWOD (\textbf{bottom}) across 4 different tasks on each. We report recall for unknown classes (U-Recall) and mAP@0.5 for known classes, separately for previously, currently, and all known objects. In each column, we highlight the \colorbox{First}{\textbf{first}}, \colorbox{Second}{\textbf{second}} and \colorbox{Third}{\textbf{third}} best results. O1O outperforms all existing OWOD methods across all tasks in terms of U-Recall except for USD~\cite{He2023ARXIV} on S-OWOD. Note that USD relies on SAM proposals which significantly contributes to its performance as can be seen from inferior results without SAM (USD \textcolor{gray}{- asf}). O1O is also consistently among the top-performing methods in terms of known mAP. Since all 80 classes are known in Task 4, U-Recall is not computed. PROB${^\dagger}$ indicate our evaluation after fixing the duplicate issue on the test set of M-OWOD.}
    \label{tab:owod_results}
\end{table*}

\boldparagraph{Comparison} 
The closest competitors in \tabref{tab:owod_results} are from DETR-based approaches, including OW-DETR~\cite{Gupta2022CVPR}, PROB~\cite{Zohar2023CVPR}, CAT~\cite{Ma2023CVPR}, and USD~\cite{He2023ARXIV}.
We encountered a bug in the evaluation of M-OWOD due to duplicated filenames in the test set. After resolving this issue\footnote{This issue might apply to other DETR-based methods. However, since the code or checkpoints are not available for them, 
except for OW-DETR, which is outperformed by PROB, 
we could not check or re-evaluate them.}, we report the improved results of PROB as  PROB${^\dagger}$.
PROB does not rely on any pseudo-labels but instead learns an objectness distribution. This allows them to achieve the best performance in terms of known AP on several tasks. However, in terms of unknown recall, PROB falls behind other approaches that use pseudo-labels such as MEPU~\cite{Fang2023ARXIV} and USD~\cite{He2023ARXIV}. Our method also uses pseudo-labels and outperforms all these methods except for USD~\cite{He2023ARXIV} on S-OWOD in terms of unknown performance while being among the top three in terms of known performance.

\boldparagraph{On the Source of Pseudo-Labels} 
To assess the effectiveness of geometric pseudo-labels, we compare our method to previous approaches that also utilize pseudo-labels from alternative sources. Specifically, we compare against MEPU~\cite{Fang2023ARXIV} which uses pseudo-labels from Object Localization Network (OLN~\cite{Kim2021RAL}). OLN, like GOOD~\cite{Huang2023ICLR}, trains the Region Proposal Network in a class-agnostic manner but relies solely on appearance cues. By leveraging geometric cues as a better source for pseudo-labels, as proposed in GOOD~\cite{Huang2023ICLR}, our method outperforms MEPU across all benchmarks. Additionally, we compare our method to USD~\cite{He2023ARXIV}, which utilizes pseudo-labels from SAM~\cite{Kirillov2023ICCV} and achieves an impressive unknown recall. Despite the advantage of USD in terms of pseudo-label source due to the internet-scale training data of SAM~\cite{Kirillov2023ICCV}, our method, leveraging geometric pseudo-labels, outperforms USD on M-OWOD and approaches its performance on S-OWOD. For additional comparison of pseudo-label sources, we trained our model using RPN proposals extracted from RGB images, predicted depth or surface normal maps, in addition to high activation areas and SAM proposals. 
Please see Supplementary for the details of this ablation.

\boldparagraph{Incremental Learning} 
The active selection strategy, proposed by PROB~\cite{Zohar2023CVPR} results in the top performance for both previously and the current known in Task 1 on the M-OWOD benchmark. We adopt a similar strategy using our proposed class conditional probability \eqref{eq:score} to select the exemplars. As more classes become known in Task 2, our method starts to catch up, eventually outperforming PROB in Task 3 and 4 when most/all classes become known. This is because our superclass representation benefits more from the variety that comes with newly introduced classes in each task by improving its class conditional probabilities, and consequently enhancing our selection of exemplars for the next tasks. 
On S-OWOD, other methods such as MEPU~\cite{Fang2023ARXIV} and CAT~\cite{Ma2023CVPR} also perform competitively, benefiting from the availability of more data to learn from in Task 1. We observe a similar trend in our model's performance in M-OWOD. As more classes become known through each task, our model improves its representation of superclasses, leading to increases in its ranking in terms of known performance.

\subsection{Ablation Study}
\label{sec:ablation}
In \tabref{tab:comp_ablation}, we ablate each component incrementally to show the effect of each on known and unknown performance. We use Task 1 of the M-OWOD benchmark for the component ablation study.

\boldparagraph{Architecture} Our baseline is Deformable DETR.
%as common choice in previous DETR-based approaches.
We first augment the model with a K+1 unknown index and supervise each unmatched query to predict this label.
This greedy strategy of predicting everything else as unknown achieves good U-Recall but leads to some loss in known AP. 
Next, we evaluate the impact of architectural improvements to Deformable DETR. While these changes significantly enhance convergence and slightly improve known AP performance, they do not yield substantial gains in unknown performance.

\boldparagraph{Geometric Pseudo-Labels} Unknown performance improves significantly when we add geometric pseudo-labels. 
For context, the previous state-of-the-art method USD reports $36.1$ for unknown recall~\cite{He2023ARXIV}. This improvement 
largely stems from SAM proposals, as evidenced by the lower recall without SAM (USD \vs USD - asf). Geometric pseudo-labels enable a notable performance jump to $50.7$, despite using the same limited training set as the detector, unlike SAM. 

\begin{figure}[t]
  \begin{minipage}{0.48\textwidth}
    \centering
    \adjustbox{max width=\textwidth}{%
      \begin{tabular}{ l cc }
        \toprule
        Metrics & U-Recall $(\uparrow)$ & mAP $(\uparrow)$  \\
        \midrule
        $1 - \text{max}\big(p\left(\bs \vert \bq\right)\big)$ & 50.7 & 72.2 \\
        $1 - \text{max}\big(p'\left(\bc \vert \bq\right)\big)$ & 52.8 & 68.2 \\
        $1 - \text{sum}\big(p'\left(\bc \vert \bq\right)\big)$ & 52.5 & 71.5 \\
        \midrule 
        $1 - \text{sum}\big(p'\left(\bc \vert \bq\right)\big)\text{+ thr.} $ & 49.8 & 72.6 \\
        \bottomrule
      \end{tabular}
    }
    \captionof{table}{\textbf{Inference Score Ablation.} We perform an ablation study on the inference score used for unknowns on S-OWOD Task 1. The first row corresponds to MSP~\cite{Hendrycks2017ICLR} applied to superclass probabilities, and the second row is MSP applied to recalibrated class probabilities $p'$~\eqref{eq:score}.
    The third row is an extended version of MSP, which uses the union of recalibrated class probabilities $p'$. As shown in the last row, we use the third with a threshold that is experimentally set on the training set.}
    \label{tab:score_ablation}
  \end{minipage}%
  \hspace{0.03\textwidth}
  \begin{minipage}{0.50\textwidth}
    \centering
    \adjustbox{max width=\textwidth}{%
      \begin{tabular}{ l c c }
        \toprule
        Components & U-Recall $(\uparrow)$ & mAP $(\uparrow)$  \\
        \midrule
        Deformable DETR & 32.6 & 59.2 \\
        + Arch. Improvements & 31.6 & 61.4  \\
        + Geo. Pseudo-Labels & 50.7 & 55.0 \\
        + Superclasses & 49.3 & 65.1 \\
        \midrule
        PROB & 28.3 & 66.4 \\
        + Arch. Improvements & 27.8 & 66.1 \\
        + Geo. Pseudo-Labels & 50.1 & 59.0 \\
        \bottomrule 
      \end{tabular}
    }
    \captionof{table}{\textbf{Component Analysis.} We perform an ablation study on Task 1 of M-OWOD by integrating our contributions one by one to the baseline, Deformable DETR, starting with architectural improvements, then geometric pseudo-labels, and finally superclasses. We also apply the same changes to PROB at the bottom part by using their objectness head instead of superclasses. }
    \label{tab:comp_ablation}
  \end{minipage}
\end{figure}

\begin{figure}[t]
  \centering
  \includegraphics[width=\linewidth]{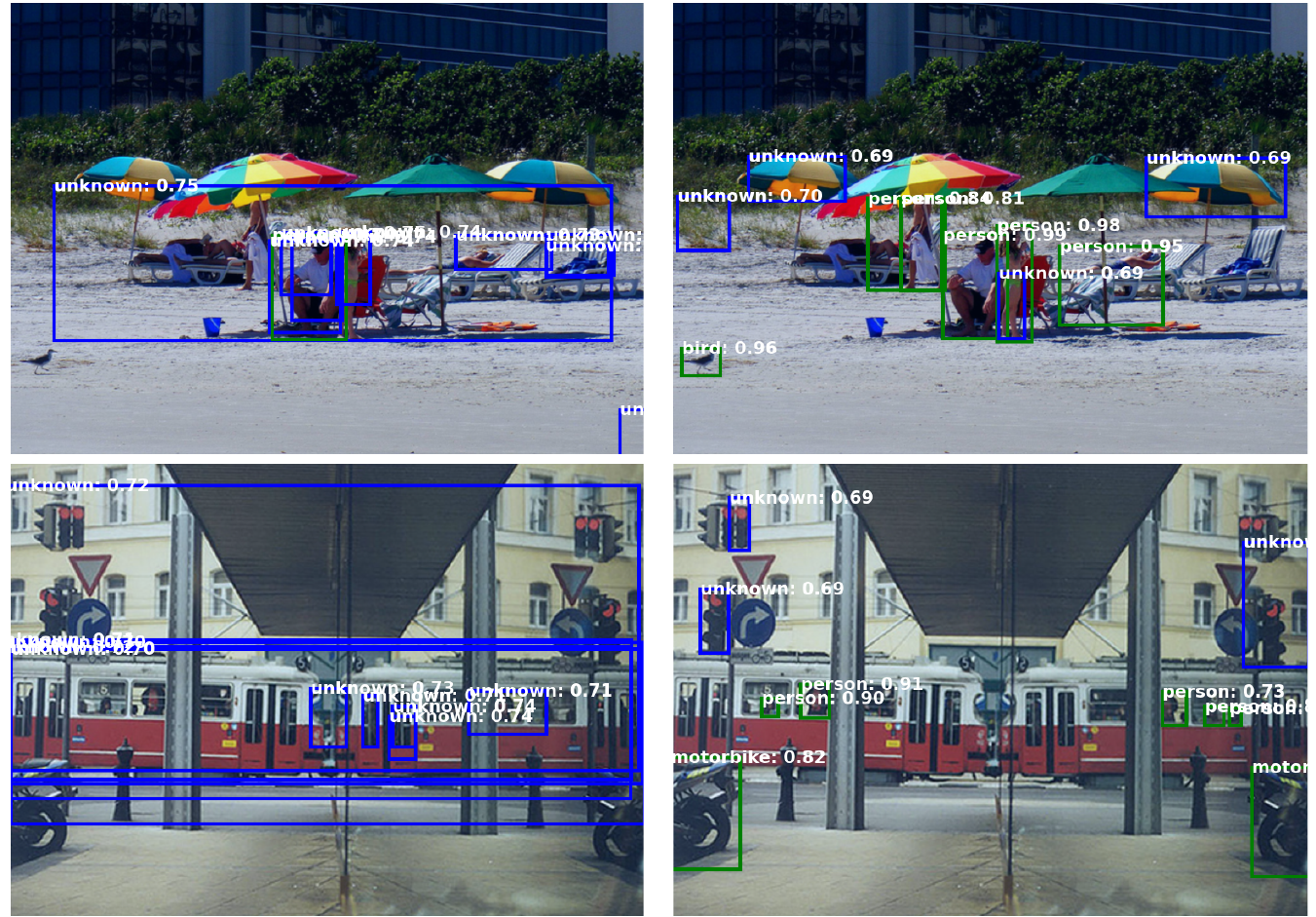}
  \caption{\textbf{Qualitative Comparison of Top-10 Predictions.} We visualize top-10 predictions for PROB~\cite{Zohar2023CVPR} (\textbf{left}) and our model O1O (\textbf{right}). We color known predictions in green and unknown in blue. Top-10 predictions of PROB are almost all unknown and duplicated over large regions, whereas our model can predict known classes like person, bird, and motorbike with high confidence and then focuses on unknown with  
  lower confidence scores, indicating a better calibration of class confidence scores. 
  }
  \label{fig:top10}
\end{figure}

\boldparagraph{Superclasses} The impressive improvement in unknown performance comes at the expense of notable degradation in known AP ($61.4 \to 55.0$), contradicting the intended goal of OWOD to preserve known performance. 
Our superclass approach addresses this by recalibrating the classification scores, resulting in a good balance between known and unknown performance. The visualization of the top-10 predictions by our model \vs PROB in \figref{fig:top10} clearly shows the benefit of this recalibration. We also test our method by varying superclass groups, demonstrating its generalization to different groupings (see Supplementary).

\boldparagraph{PROB with Benefits} To further clarify our contribution with superclasses, we apply the same changes to PROB at the bottom part of \tabref{tab:comp_ablation}. Compared to the greedy Deformable DETR baseline, PROB's objectness probability helps maintain a high known performance, by ranking background regions lower in the confidence scores. Applying the same architectural improvements to PROB does not affect its performance. Incorporating geometric pseudo-labels improves the unknown performance of PROB by almost doubling the unknown recall ($27.8 \to 50.1$) but also negatively impacts the known performance ($-7.1$ in mAP), similar to what we observed with our method. PROB's approach, which involves learning a single distribution to represent objectness, fails to mitigate the degradation in known performance. This underscores the importance of shaping the representation space with superclasses (\figref{fig:feat_space}).

\boldparagraph{Unknown Scoring Function} We define our unknown probability based on the known class confidence scores \eqref{eq:unk_score}. In \tabref{tab:score_ablation}, we report the performance of alternative scoring functions on S-OWOD, Task 1. For a confident known prediction, the maximum softmax probability (MSP)~\cite{Hendrycks2017ICLR} becomes high, and the inverse becomes low, so we can obtain the unknown probability by simply subtracting it from 1. We apply MSP on superclass probabilities alone in the first row, and on recalibrated class probabilities \eqref{eq:score} in the second row. In the third row, we extend the max in MSP to the union of recalibrated class probabilities. This indicates the combined belief of the model about a prediction to belong to a known class. 
Depending on the type of scoring function, there is a trade-off between known and unknown performances.
We found the third approach to have the best balance between U-Recall and known AP. Furthermore, we calculate a threshold on a subset of the training set such that $95\%$ of the known instances would be classified correctly. This thresholding helps us control the trade-off between known and unknown slightly better, without violating any principles.
\section{Conclusion and Future Work}
\label{sec:conclusion}
In this work, we build on DETR-based object detection models by better utilizing unmatched queries for unknown object detection. We first integrate pseudo-labels from a class-agnostic region proposal network based on geometrical cues, \ie surface normal maps. Integrating
pseudo-labels increases unknown recall significantly. 
However, it requires better modeling of known classes to handle the increased confusion caused by noisy pseudo-labels. 
To address this, we introduce a prior on the queries using superclass groupings in addition to classification scores. Superclass supervision helps recalibrate the scores of known classes, effectively reducing confusion due to pseudo-labels. Our approach results in improved known performance with the help of superclasses while being among the top-performing methods in terms of unknown recall. 
We use a one-level hierarchy of superclasses, 
% but there might be potential gains from exploring more fine-grained hierarchies, especially for large categories like vehicles. 
but also show that O1O generalizes across different superclass groupings. %as long as the inner class dynamics are addressed properly.
Our method relies on geometric cues for localizing unknowns which poses a challenge when these cues cannot be reliably predicted. Additionally, balancing noise reduction with accurate localization remains a critical challenge.

\begin{credits}
\subsubsection{\ackname} 
We thank João F. Henriques and Nazir Nayal for their valuable feedback. 
% during the project. 
This project is co-funded by KUIS AI, Royal Society Newton Fund Advanced Fellowship (NAF\textbackslash R\textbackslash 2202237), and the European Union (ERC, ENSURE, 101116486). Views and opinions expressed are those of the author(s) only and do not necessarily reflect those of the European Union or the European Research Council. Neither the European Union nor the granting authority can be held responsible for them.

% \subsubsection{\discintname}
% It is now necessary to declare any competing interests or to specifically
% state that the authors have no competing interests. Please place the
% statement with a bold run-in heading in small font size beneath the
% (optional) acknowledgments\footnote{If EquinOCS, our proceedings submission
% system, is used, then the disclaimer can be provided directly in the system.},
% for example: The authors have no competing interests to declare that are
% relevant to the content of this article. Or: Author A has received research
% grants from Company W. Author B has received a speaker honorarium from
% Company X and owns stock in Company Y. Author C is a member of committee Z.
\end{credits}
%
% ---- Bibliography ----
%
% BibTeX users should specify bibliography style 'splncs04'.
% References will then be sorted and formatted in the correct style.
%
\bibliographystyle{splncs04}
\bibliography{main}

\begin{thebibliography}{10}
\providecommand{\url}[1]{\texttt{#1}}
\providecommand{\urlprefix}{URL }
\providecommand{\doi}[1]{https://doi.org/#1}

\bibitem{Bendale2015CVPR}
Bendale, A., Boult, T.E.: Towards open world recognition. In: CVPR (2015)

\bibitem{Carion2020ECCV}
Carion, N., Massa, F., Synnaeve, G., Usunier, N., Kirillov, A., Zagoruyko, S.: End-to-end object detection with transformers. In: ECCV (2020)

\bibitem{Caron2021ICCV}
Caron, M., Touvron, H., Misra, I., J'egou, H., Mairal, J., Bojanowski, P., Joulin, A.: Emerging properties in self-supervised vision transformers. In: ICCV (2021)

\bibitem{Deng2009CVPR}
Deng, J., Dong, W., Socher, R., Li, L.J., Li, K., Fei-Fei, L.: {ImageNet:} a large-scale hierarchical image database. In: CVPR (2009)

\bibitem{Dhamija2020WACV}
Dhamija, A.R., G{\"u}nther, M., Ventura, J., Boult, T.E.: The overlooked elephant of object detection: Open set (2020)

\bibitem{Du2022NeurIPS}
Du, X., Gozum, G., Ming, Y., Li, Y.: {SIREN}: Shaping representations for detecting out-of-distribution objects. In: NeurIPS (2022)

\bibitem{du2023superdisco}
Du, Y., Shen, J., Zhen, X., Snoek, C.G.: Superdisco: Super-class discovery improves visual recognition for the long-tail. In: CVPR (2023)

\bibitem{Eftekhar2021ICCV}
Eftekhar, A., Sax, A., Malik, J., Zamir, A.: Omnidata: A scalable pipeline for making multi-task mid-level vision datasets from 3d scans. In: ICCV (2021)

\bibitem{Fang2023ARXIV}
Fang, R., Pang, G., Zhou, L., Bai, X., Zheng, J.: Unsupervised recognition of unknown objects for open-world object detection  (2023)

\bibitem{Girshick2015ICCV}
Girshick, R.: Fast r-cnn. In: ICCV (2015)

\bibitem{Girshick2013CVPR}
Girshick, R.B., Donahue, J., Darrell, T., Malik, J.: Rich feature hierarchies for accurate object detection and semantic segmentation. In: CVPR (2013)

\bibitem{Gupta2022CVPR}
Gupta, A., Narayan, S., Joseph, K., Khan, S., Khan, F.S., Shah, M.: {OW-DETR:} open-world detection transformer. In: CVPR (2022)

\bibitem{Hall2018WACV}
Hall, D., Dayoub, F., Skinner, J., Zhang, H., Miller, D., Corke, P., Carneiro, G., Angelova, A., S{\"u}nderhauf, N.: Probabilistic object detection: Definition and evaluation (2018)

\bibitem{Han2022CVPR}
Han, J., Ren, Y., Ding, J., Pan, X., Yan, K., Xia, G.: Expanding low-density latent regions for open-set object detection. In: CVPR (2022)

\bibitem{He2015CVPR}
He, K., Zhang, X., Ren, S., Sun, J.: Deep residual learning for image recognition. In: CVPR (2015)

\bibitem{He2023ARXIV}
He, Y., Chen, W., Tan, Y., Wang, S.: {USD:} unknown sensitive detector empowered by decoupled objectness and segment anything model  \textbf{2306.02275} (2023)

\bibitem{Hebart2020revealing}
Hebart, M.N., Zheng, C.Y., Pereira, F., Baker, C.I.: Revealing the multidimensional mental representations of natural objects underlying human similarity judgements. Nature human behaviour  \textbf{4}(11),  1173--1185 (2020)

\bibitem{Hendrycks2017ICLR}
Hendrycks, D., Gimpel, K.: A baseline for detecting misclassified and out-of-distribution examples in neural networks. In: ICLR (2017)

\bibitem{Huang2023ICLR}
Huang, H., Geiger, A., Zhang, D.: {GOOD}: Exploring geometric cues for detecting objects in an open world. In: ICLR (2023)

\bibitem{Joseph2021CVPR}
Joseph, K.J., Khan, S., Khan, F.S., Balasubramanian, V.N.: Towards open world object detection. In: CVPR (2021)

\bibitem{Kim2021RAL}
Kim, D., Lin, T.Y., Angelova, A., Kweon, I.S., Kuo, W.: Learning open-world object proposals without learning to classify  \textbf{PP}, ~1--1 (2021)

\bibitem{Kirillov2023ICCV}
Kirillov, A., Mintun, E., Ravi, N., Mao, H., Rolland, C., Gustafson, L., Xiao, T., Whitehead, S., Berg, A.C., Lo, W.Y., Dollár, P., Girshick, R.: Segment anything. In: ICCV (2023)

\bibitem{Li2022CVPR}
Li, F., Zhang, H., Liu, S., Guo, J., Ni, L.M., Zhang, L.: {DN-DETR}: Accelerate detr training by introducing query denoising. In: CVPR (2022)

\bibitem{li2021sgnet}
Li, K., Wang, N.Y., Yang, Y., Wang, G.: Sgnet: A super-class guided network for image classification and object detection. In: CRV (2021)

\bibitem{Lin2017ICCV}
Lin, T.Y., Goyal, P., Girshick, R., He, K., Doll{\'a}r, P.: Focal loss for dense object detection. In: ICCV (2017)

\bibitem{Liu2021ICLR}
Liu, S., Li, F., Zhang, H., Yang, X., Qi, X., Su, H., Zhu, J., Zhang, L.: Dab-detr: Dynamic anchor boxes are better queries for detr. In: ICLR (2021)

\bibitem{Ma2023CVPR}
Ma, S., Wang, Y., Fan, J., yu~Wei, Y., Li, T.H., Liu, H., Lv, F.: {CAT:} localization and identification cascade detection transformer for open-world object detection. In: CVPR (2023)

\bibitem{Maaz2021ECCV}
Maaz, M., Rasheed, H.A., Khan, S., Khan, F.S., Anwer, R.M., Yang, M.: Class-agnostic object detection with multi-modal transformer. In: ECCV (2021)

\bibitem{Maaz2021ARXIV}
Maaz, M., Rasheed, H.A., Khan, S.H., Khan, F.S., Anwer, R.M., Yang, M.H.: Multi-modal transformers excel at class-agnostic object detection  (2021)

\bibitem{Mccloskey1989}
McCloskey, M., Cohen, N.J.: Catastrophic interference in connectionist networks: The sequential learning problem. In: Psychology of learning and motivation, vol.~24, pp. 109--165. Elsevier (1989)

\bibitem{Miller2018ICRA}
Miller, D., Dayoub, F., Milford, M., S{\"u}nderhauf, N.: Evaluating merging strategies for sampling-based uncertainty techniques in object detection (2018)

\bibitem{Miller2017ICRA}
Miller, D., Nicholson, L., Dayoub, F., S{\"u}nderhauf, N.: Dropout sampling for robust object detection in open-set conditions (2017)

\bibitem{Miller2021RAL}
Miller, D., Sunderhauf, N., Milford, M., Dayoub, F.: Uncertainty for identifying open-set errors in visual object detection  \textbf{7},  215--222 (2021)

\bibitem{Ranftl2021ICCV}
Ranftl, R., Bochkovskiy, A., Koltun, V.: Vision transformers for dense prediction. In: ICCV (2021)

\bibitem{Ren2015PAMI}
Ren, S., He, K., Girshick, R.B., Sun, J.: Faster {R-CNN:} towards real-time object detection with region proposal networks. IEEE TPAMI  \textbf{39},  1137--1149 (2015)

\bibitem{Saito2022ECCV}
Saito, K., Hu, P., Darrell, T., Saenko, K.: Learning to detect every thing in an open world. In: ECCV (2022)

\bibitem{Uijlings2013IJCV}
Uijlings, J.R.R., van~de Sande, K.E.A., Gevers, T., Smeulders, A.W.M.: Selective search for object recognition. IJCV  \textbf{104},  154 -- 171 (2013)

\bibitem{Wang2023ICCV}
Wang, Y., Yue, Z., Hua, X.S., Zhang, H.: Random boxes are open-world object detectors. In: ICCV (2023)

\bibitem{Wu2022HCMA@MM}
Wu, Y., Zhao, X., Ma, Y., Wang, D., Liu, X.: Two-branch objectness-centric open world detection. Proceedings of the 3rd International Workshop on Human-Centric Multimedia Analysis  (2022)

\bibitem{Wu2022ECCV}
Wu, Z., Lu, Y., Chen, X., Wu, Z., Kang, L., Yu, J.: {UC-OWOD:} unknown-classified open world object detection. In: ECCV (2022)

\bibitem{Yu2022ICIP}
Yu, J., Ma, L., Li, Z., Peng, Y., Xie, S.: Open-world object detection via discriminative class prototype learning. In: ICIP (2022)

\bibitem{Zhao2022ECCV}
Zhao, S., Zhang, Z., Schulter, S., Zhao, L., B.G, V.K., Stathopoulos, A., Chandraker, M., Metaxas, D.N.: Exploiting unlabeled data with vision and language models for object detection. In: ECCV (2022)

\bibitem{Zheng2022CVPRW}
Zheng, J., Li, W., Hong, J., Petersson, L., Barnes, N.: Towards open-set object detection and discovery. 2022 IEEE/CVF Conference on Computer Vision and Pattern Recognition Workshops (CVPRW)  (2022)

\bibitem{zhou2018deep}
Zhou, Y., Hu, Q., Wang, Y.: Deep super-class learning for long-tail distributed image classification. PR  \textbf{80},  118--128 (2018)

\bibitem{Zhu2021ICLR}
Zhu, X., Su, W., Lu, L., Li, B., Wang, X., Dai, J.: Deformable {DETR}: Deformable transformers for end-to-end object detection. In: ICLR (2021)

\bibitem{Zohar2023CVPR}
Zohar, O., Wang, K.C., Yeung, S.: {PROB:} probabilistic objectness for open world object detection. In: CVPR (2023)

\end{thebibliography}
%

% \clearpage

\title{Supplementary Material for "O1O: Grouping of Known Classes to Identify Unknown Objects as Odd-One-Out"} 
%
% If the paper title is too long for the running head, you can set
% an abbreviated paper title here
\titlerunning{O1O}
%

% TODO FINAL: Replace with your author list. 
% Include the authors' OCRID for the camera-ready version, if at all possible.
\author{Mısra Yavuz\inst{1}\orcidlink{0009-0004-9517-2506} \and
Fatma Güney\inst{1}\orcidlink{0000-0002-0358-983X} 
% \and Third Author\inst{3}\orcidlink{2222--3333-4444-5555}
}
\authorrunning{M. Yavuz et al.}
% First names are abbreviated in the running head.
% If there are more than two authors, 'et al.' is used.
%
\institute{
Department of Computer Engineering, Koç University \\
KUIS AI Center\\
\email{\{myavuz21,fguney\}@ku.edu.tr}
}

\maketitle              % typeset the header of the contribution

\begin{abstract}
This supplementary document provides our superclass formulation (\secref{sec:super_class}), implementation details for our method (\secref{sec:implementation_details}), additional ablations for superclass groups as well as source and number of pseudo-labels (\secref{sec:more_ablations}), and qualitative examples including some failure cases and comparison to previous methods (\secref{sec:qualitative}).

\end{abstract}  

\section{Superclass Grouping}
\label{sec:super_class}
\begin{figure}[b]
\centering
\begin{minipage}{0.95\textwidth}
    \centering
    \adjustbox{max width=\textwidth}{%
        \begin{tabular}{c c c}
            \toprule
            \hspace{.1in} Benchmark \hspace{.1in} & \hspace{.1in} Task \hspace{.1in} & Superclasses Used During Training \\
            \midrule
            \multirow{4}{*}{S-OWOD} & 1 & animal, person, vehicle \\
            & 2 & accessory, appliance, furniture, outdoor \\
            & 3 & food, sports \\
            & 4 & electronic, indoor, kitchen \\
            \midrule
            \multirow{4}{*}{M-OWOD} & 1 & \hspace{.1in} \underline{animal}, \underline{electronic}, \underline{furniture}, \underline{kitchen}, person, \underline{vehicle}  \hspace{.1in} \\
            & 2 & accessory, \underline{animal}, appliance, outdoor, \underline{vehicle} \\
            & 3 & food, sports \\
            & 4 & \underline{electronic}, \underline{furniture}, indoor, \underline{kitchen} \\
            \bottomrule
        \end{tabular}}
    \captionof{table}{\textbf{Superclass Separation Across Tasks.} }
    \label{tab:sup_info}
\end{minipage}
\end{figure}

The OWOD benchmarks are based on the set of COCO classes, which follow a superclass hierarchy specified in the official COCO annotation files.
Following the same hierarchy and the benchmark splits, superclasses introduced at each task for both S-OWOD and M-OWOD benchmarks can be found in \tabref{tab:sup_info}. 
Since S-OWOD tasks are designed to have perfect superclass separation, each superclass introduced in a task is learned from scratch and to completion. 
On the other hand, on M-OWOD, a task might have only some of the classes belonging to a certain superclass in the current training set. 
For example, in M-OWOD Task 1, only \textit{bird}, \textit{cat}, \textit{cow}, \textit{dog}, \textit{horse}, and \textit{sheep} classes are learned from the animal superclass; while \textit{elephant}, \textit{bear}, \textit{zebra}, and \textit{giraffe} are learned in Task 2. 
Regardless, if any class of a superclass is present, we introduce it at the current task. 
Unseen classes are still considered unknown, following the benchmark rules.
As the unseen classes become available in the continual learning procedure, we continue training the specific weights of the partially introduced superclasses. 
We underline partially learned superclasses in \tabref{tab:sup_info}. 
Although partially learning the superclass representations hurt the odd-one-out scoring in theory, our experimental results show that our method can still generalize to such a setting.

\section{Implementation Details}
\label{sec:implementation_details}
Our pipeline can be considered as a two-step approach: first, we train the RPN to extract pseudo-labels, then we train our open-world detection model.

\boldparagraph{Pseudo-labels Extraction Step} We train the RPN with surface normal maps estimated by the DPT-Hybrid model~\cite{Ranftl2021ICCV} from the Omnidata repository~\cite{Eftekhar2021ICCV} as in GOOD~\cite{Huang2023ICLR} and store the proposals. 
We train the RPN of each task from scratch to obey the benchmark rules and not use any unknown instances for training. 
We select the pseudo-labels by thresholding the RPN's confidence score with 0.5 and then merge them with real targets by applying Non-Maximum Suppression with an IoU threshold of 0.5. 

\boldparagraph{OWOD Step} Our detection model is based on the DN-DAB-Deformable DETR~\cite{Liu2021ICLR,Li2022CVPR} architecture, with the same feature extraction backbone as previous work: ResNet-50 FPN~\cite{He2015CVPR} pre-trained on ImageNet~\cite{Deng2009CVPR} in a self-supervised manner~\cite{Caron2021ICCV}.
We decompose our loss function into localization and classification. 
The localization losses are applied to queries matched to both knowns and pseudo-unknowns, while classification losses are only applied to queries matched to known targets. 
We use loss coefficients 5 for bounding box loss, 2 for giou loss, 2
 for per-class classification loss, and 2 for superclass classification loss. 
We follow the default hyperparameters of DN-DAB-Deformable DETR for the denoising tasks. In addition, we add a denoising task for superclass classification, using the same hyperparameters as the per-class denoising loss, to randomly flip the ground truth labels for noised queries and teach the model to correct them.

\boldparagraph{M-OWOD Bug Fix} 
\begin{table}[b]
    \centering
    \begin{tabularx}{0.6\textwidth}{l *{2}{>{\centering\arraybackslash}X}}
        \toprule
        Method & U-Recall $(\uparrow)$ & mAP $(\uparrow)$  \\
        \midrule
        OW-DETR$^{\dagger}$ & 7.6 $\to$ 10.1 & 58.8 $\to$ 65.6 \\
        PROB & 19.4 $\to$ 28.3 & 59.5 $\to$ 66.4 \\
        O1O & 33.4 $\to$ 49.3 & 58.4 $\to$ 65.1 \\
        \bottomrule 
      \end{tabularx}
    \caption{\textbf{Performance Improvements after M-OWOD Bug Fix} 
    }
    \label{tab:mowod_bug}
\end{table}

We found duplicate image IDs in the test set of M-OWOD, which can affect the performance of DETR-based methods based on how the matching operation is done during evaluation. 
We detected that OW-DETR and PROB were affected by this bug. 
The codes of more recent DETR-based methods, CAT and USD, were not available to check. 
In \tabref{tab:mowod_bug}, we report performance changes of OW-DETR and PROB on Task 1 after fixing the bug. 
We trained OW-DETR from scratch for Task 1 (denoted with $^{\dagger}$), because we couldn't reproduce their results with the checkpoints provided. 
Therefore, we can only report the performance change in Task 1. 
PROB's results on all tasks, with and without the fix, are also reported in Table 1 of the main paper. 
O1O's performance, with and without the fix, are consistent. 
We achieve significantly higher Unknown Recall while maintaining competitive performance in Known mAP.
Compared to PROB, we obtain +74.2\% increase in Unknown Recall with only -2.0\% decrease in Known mAP.

\section{Additional Ablations}
\label{sec:more_ablations}

\subsection{Varying Superclass Groups}
\label{sec:super_groups}

\begin{table*}[h]
    \centering
    
    \begin{subtable}[]{\linewidth}
        \centering

        \begin{tabular}{l c c c c}
            \toprule
            \textbf{Group ID} & \textbf{Superclasses} \\
            \midrule
            
            A & living beings, objects \\
            \midrule
            
            \multirow{2}{*}{B} & pets, wild animals, land vehicles, air/water vehicles, \\ 
            & seating furniture, household items, person \\ 
            \midrule
            
            \multirow{2}{*}{C} & pets, farm animals, wild animals,  bikes (2-wheel), land vehicles (4-wheel), \\
            & air/water vehicles, seating furniture, electronics, household items, person \\ 
            \midrule        

            D (Default) & animals, vehicles, person, furniture, electronics, kitchen \\

            \bottomrule
        \end{tabular}
        \caption{\textbf{M-OWOD}}
    \end{subtable}

    \begin{subtable}[]{\linewidth}
        \centering

        \begin{tabular}{l c c c c}  
            \toprule
            \textbf{Group ID} & \textbf{Superclasses} \\
            \midrule
            A & living beings, objects \\
            \midrule        
            
            B & domestic animals, wild animals, person, land vehicles, air/water vehicles \\
            \midrule        
            
            \multirow{2}{*}{C} & large animals, medium animals, small animals, person \\
            & bikes (2-wheel), land vehicles (4-wheel), air vehicles, water vehicles \\ 
            \midrule        

            D (Default) & animals, person, vehicles \\
            
            \bottomrule 
        \end{tabular}
        \caption{\textbf{S-OWOD}}
    \end{subtable}
    \caption{\textbf{Different Groupings of Classes into Superclasses}}
    \label{tab:diff_groups}
\end{table*}

To evaluate the robustness of our method to different superclass groups, we explore broader and narrower sets of superclasses than the default grouping (D) used in the main paper in \tabref{tab:diff_groups}.
First, in A, we combined all non-living things into a single objects class and grouped animals and humans as living beings.
Conversely, we explored more fine-grained groupings in B and C by dividing animals into categories like pets, farm animals, domestic or wild animals, or by size into large, medium, and small. Additionally, we organized vehicles based on the number of wheels and their typical usage. 
The default (D) groups for each benchmark are formed using the official superclass labels in COCO annotations. 
We used the default groups in our main results in the paper to comply with recognized standards and minimize subjective biases.

\begin{table}[ht]
    \centering
    \begin{tabular}{ l c c c c}
        \toprule
        \multirow{2}{*}{\textbf{Group ID}} & \multicolumn{2}{c}{\textbf{M-OWOD}} & \multicolumn{2}{c}{\textbf{S-OWOD}} \\
        \cmidrule(r){2-3}
        \cmidrule(r){4-5}
        & \textbf{U-Recall} & \textbf{mAP} &  \textbf{U-Recall} & \textbf{mAP}  \\
        \midrule
        A & 40.0 & 64.9 & 48.2 & 72.1 \\
        B & 48.5 & 65.0 & 51.3 & 72.5 \\
        C & 48.9 & 64.8 & 51.4 & 70.8 \\
        D (Default) & 49.3 & 65.1 & 49.8 & 72.6 \\
        \bottomrule 
      \end{tabular}
    \caption{\textbf{Results by Varying Superclass Groups}}
    \label{tab:results}
\end{table}

We report the results of this ablation on Task 1 of both benchmarks in \tabref{tab:results}. Group A consists of living beings and objects. On M-OWOD, forcing various objects such as furniture, electronics, kitchen items, and vehicles into a single group leads to an oversimplification. This results in a generic distribution to which any class can belong including unknowns, hence the drop in unknown recall. 
For S-OWOD, since objects only consist of vehicles, the setup is very close to default, with the only difference being the merging of animal and person classes. As a result, the performance drop is not as significant. Note that animal and person classes are similar enough that the `living beings' distribution does not unintentionally cover representations of other classes.

For B, the results are similar to the default case. As superclasses narrow down in Group C, known performance drops while unknown performance increases slightly. This can be attributed to unknowns standing out more as odd-one-outs as superclass representation groups become smaller in the feature space. Overall, these results confirm that our method performs well across different superclass groups with varying granularity, supporting the robustness of our approach.

\subsection{Source of Pseudo-labels}

To analyze the strength of geometric pseudo-labels with our method, we performed an additional study to compare them to RGB-based proposals. 
We trained the RPN module using three sources of input: RGB images, predicted depth maps, and predicted surface normal maps.
For clarity, the detection model always receives the RGB images as input. Here, we only modify the input to the RPN module. 
As shown in \tabref{tab:pseudo_ablation}, while the Unknown Recall of RGB and surface normals are close on the M-OWOD benchmark, the difference becomes visible in S-OWOD. The reason behind is the effect of dataset size. The M-OWOD Task 1 dataset is approximately one-sixth the size of the S-OWOD Task 1 dataset. Hence, any use of extra supervision helps the model to improve its recall significantly. However, in the abundance of training instances, such as in S-OWOD, the detection model already leverages the full spectrum of information available from RGB input. Therefore, when surface normals are used, it outperforms the RGB variant by +4.2 in Unknown Recall with a +0.7 better Known mAP. In both benchmarks, RGB pseudo-labels cause the most confusion, hurting the known mAP considerably. 

\begin{table}[t]
    \centering
    \begin{tabularx}{\textwidth}{l *{4}{>{\centering\arraybackslash}X}}
        \toprule
        \multirow{2}{*}{Source} & \multicolumn{2}{c}{M-OWOD} & \multicolumn{2}{c}{S-OWOD} \\
        \cmidrule(r){2-3}
        \cmidrule(r){4-5}
        & U-Recall $(\uparrow)$ & mAP $(\uparrow)$ &  U-Recall $(\uparrow)$ & mAP $(\uparrow)$ \\
        \midrule
        RGB & 50.5 $\to$ 51.5 & 53.2 $\to$ 62.8 &  48.7 & 70.3 \\
        Depth & 46.8 $\to$ 48.0 & 54.9 $\to$ 63.0 & 45.9 & 72.6 \\
        Normals & 50.7  $\to$ 51.6 & 55.0 $\to$ 63.6 & 52.9 & 71.0 \\
        \bottomrule 
      \end{tabularx}
    \caption{\textbf{Varying the Source of Pseudo-Labels}}
    \label{tab:pseudo_ablation}
\end{table}

Depth alone has the poorest performance among the three methods in terms of unknown performance. Due to the tradeoff between known and unknown performance, this leaves room for known performance to be relatively higher.
Overall, surface normals yield the highest unknown recall on both benchmarks and the least negative impact on known performance.

Furthermore, we performed additional experiments without the superclass component to showcase the improvements gained by using them (trained without $\to$ with). We used the M-OWOD benchmark since the smaller dataset size allows for faster experiments. The results confirm that superclasses generalize to pseudo-labels from different sources, and shaping the representation helps both Unknown Recall and Known mAP across all three settings.

\subsection{Number of Pseudo-labels}
After deciding which source of pseudo-labels to use, the next question is how many to use. 
We conducted experiments with different numbers of pseudo-labels on M-OWOD Task 1, utilizing surface normals as the source. 
We implement the number of pseudo-labels as a constraint after the NMS module and confidence thresholding. 
This means that after removing overlapping or low-confidence boxes, we select the top-$k$ boxes, where $k$ is the number of known ground truth targets plus the number of pseudo-labels.

\begin{figure}[h]
  \centering
  \includegraphics[width=\linewidth]{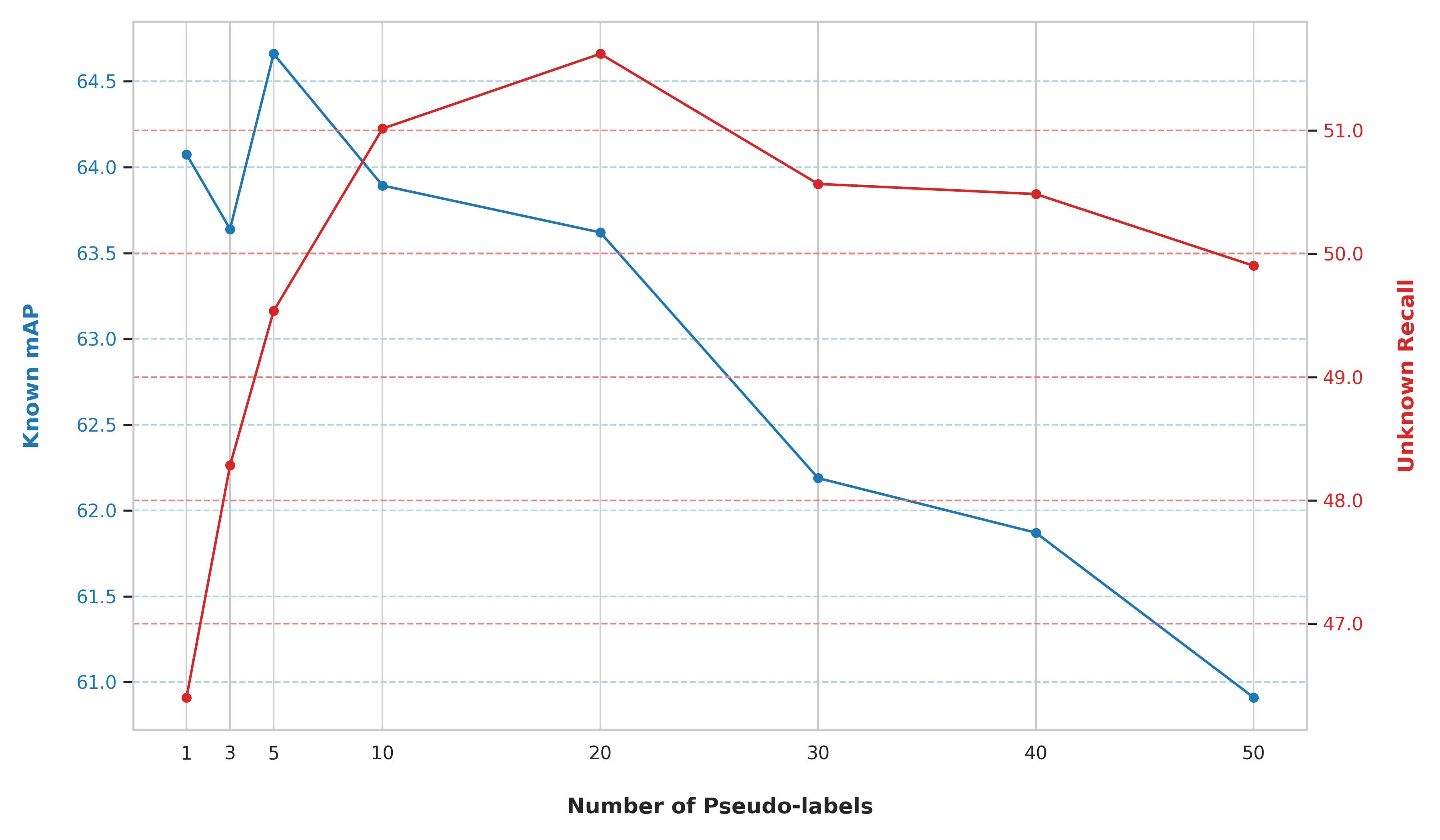}
  \caption{\textbf{Number of Pseudo-labels Ablation.} %We study the effect of utilizing different numbers of pseudo-labels on Task 1 of M-OWOD, using surface normals as the source. Unknown Recall shows a clear improvement until 20 labels, after which it starts to decrease. Known mAP remains within a range until 5 labels, then shows a clear decrease trend, particularly after surpassing 20 labels. Optimal performance appears to be attained with 20 pseudo-labels, balancing accuracy, and noise. 
  }
  \label{fig:num_unk}
\end{figure}

We report how Known mAP values evolve against Unknown Recall in \figref{fig:num_unk}. 
We start with small numbers such as 1, 3, and 5, which are typically used in previous work \cite{Joseph2021CVPR,Gupta2022CVPR,Ma2023CVPR}, and increase the number up to half of total number of queries, which is 100. 
% It is expected to see a trade-off between known and unknown performance, which is confirmed by our experiments. 
Our experiment results demonstrate the trade-off between known and unknown performance. 
As the number of pseudo-labels increases, Unknown Recall shows a clear improvement until the number 20. 
Known mAP has an unstable response to the number of pseudo-labels until the number 5, remaining in a range, after which it exhibits a clear decreasing trend, as expected.

After the number 20, both Known mAP and Unknown Recall decline because the noisy and inaccurate pseudo-supervision dominates the signals coming from the actual ground truth supervision. 
As the equilibrium point, we use 20 pseudo-labels per image in our method.
This is not implemented as a hard constraint, meaning if there are fewer than 20 pseudo-boxes with a confidence score greater than our confidence threshold (0.5), we use exactly how many there are.

\section{Qualitative Examples}
\label{sec:qualitative}
\begin{figure}
  \centering
  \includegraphics[width=\linewidth]{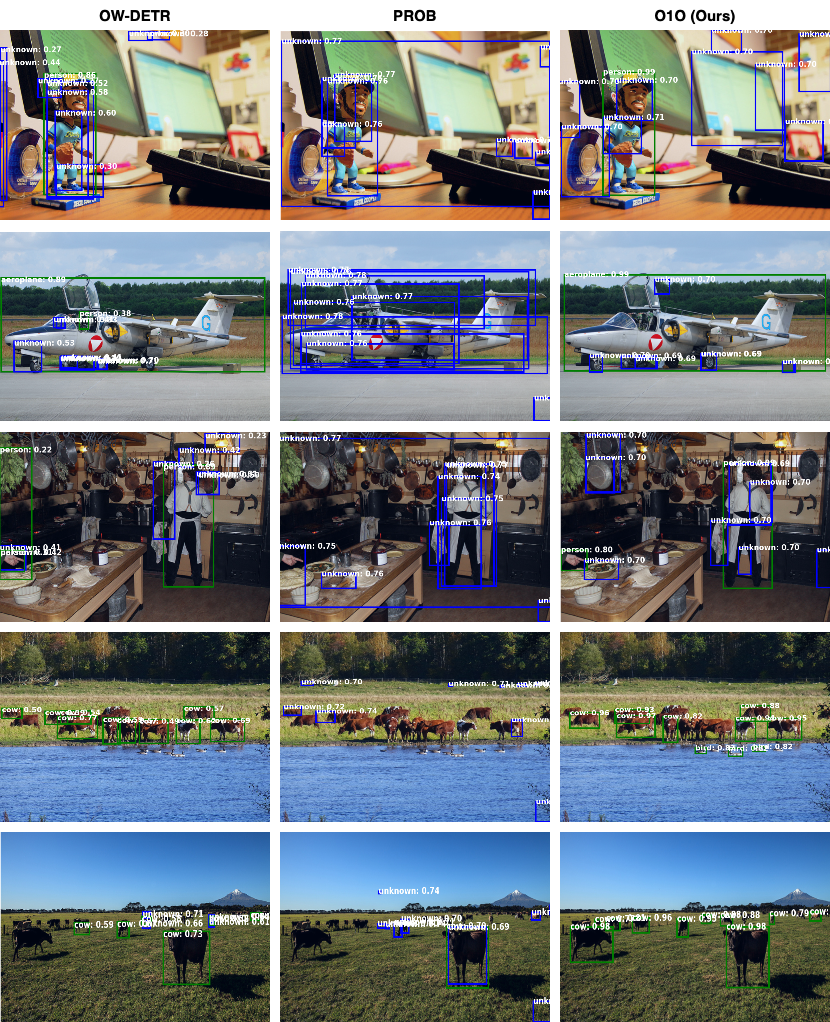}
  \caption{\textbf{Qualitative Comparison on S-OWOD.} Visualizations of top-10 proposals of OW-DETR, PROB, and O1O(Ours).}
  \label{fig:s_1_good}
\end{figure}

\begin{figure}
  \centering
  \includegraphics[width=\linewidth]{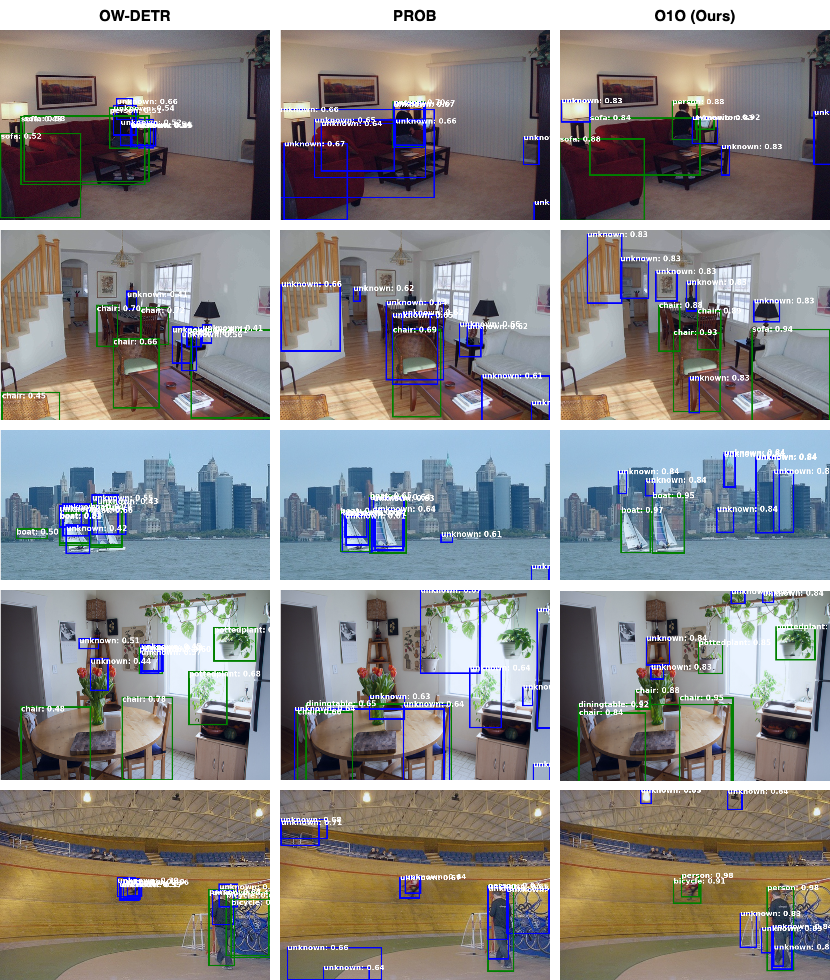}
  \caption{\textbf{Qualitative Comparison on M-OWOD.} Visualizations of top-10 proposals of OW-DETR, PROB, and O1O(Ours).}
  \label{fig:m_1_good}
\end{figure}

We visualize the top-10 proposals of OW-DETR~\cite{Gupta2022CVPR}, PROB~\cite{Zohar2023CVPR} and our model O1O in \figref{fig:s_1_good} for S-OWOD and in \figref{fig:m_1_good} for M-OWOD. 
In most cases, OW-DETR and our method have a better ranking than PROB, prioritizing known objects with higher confidence and representing unknown predictions with lower scores. 
As shown in \figref{fig:s_1_good}, our model can locate the monitor, frame and tape precisely in the first row, can detect the box on the ground and and the tires of the plane as separate objects, and finds the pans and the towel in the third row, while previous work failed to do so. 
Some detected unknowns are not even annotated in the official COCO dataset, therefore not contributing to the Unknown Recall metric, such as tape and towel. 
Lastly, without any unknowns present, such as the last two rows, our method can label known objects without any confusion and does not produce unnecessary unknown predictions with high confidence scores. 
Our method performs similarly on M-OWOD as illustrated in \figref{fig:m_1_good}. It can precisely detect the lamps, frames, buildings, or small objects in the background while maintaining a meaningful ranking between known and unknown predictions.

\begin{figure}
  \centering
  \includegraphics[width=0.99\linewidth]{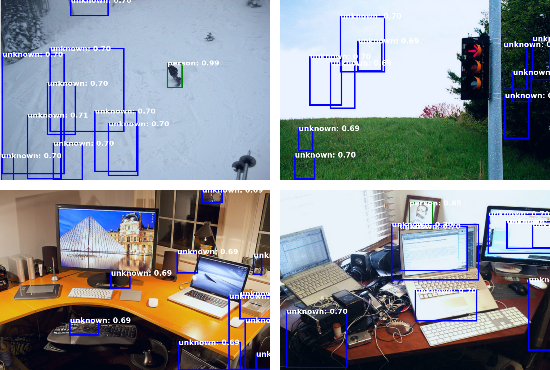}
  \caption{\textbf{Failure cases on S-OWOD.} Due to reliance on estimated surface normals, our model may generate redundant pseudo-labels in areas with local texture differences (\textbf{top}) or struggle to detect objects in crowded scenes (\textbf{bottom}).}
  \label{fig:s_1_bad}
\end{figure}

\begin{figure}
  \centering
  \includegraphics[width=0.99\linewidth]{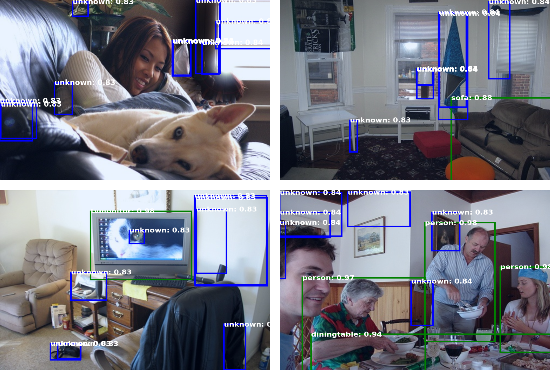}
  \caption{\textbf{Failure cases on M-OWOD.} Examples of ranking redundant unknown predictions higher than some known instances.}
  \label{fig:m_1_bad}
\end{figure}

\boldparagraph{Failure Cases} Since our pseudo-labels depend on predicted surface normals, O1O is vulnerable to cases when surface normals cannot be estimated accurately. One failure mode is when textual differences in a large background, like the sky or floor, cause relative differences in local normal values. This encourages the model to predict objects in such regions, as in the first row of \figref{fig:s_1_bad}. Another failure mode occurs in complex indoor scenes with many objects, sometimes because the objects have flat shapes, such as keyboards, or because of crowded regions with entangled objects, as in the second row of \figref{fig:s_1_bad}. 
Our model occasionally ranks blank regions that do not correspond to real objects higher than known classes, such as the person and dog in the upper left, the table in the upper right, the sofa/chair in the lower left, or the person in the lower right of \figref{fig:m_1_bad}.

\begin{figure}
  \centering
  \includegraphics[width=0.99\linewidth]{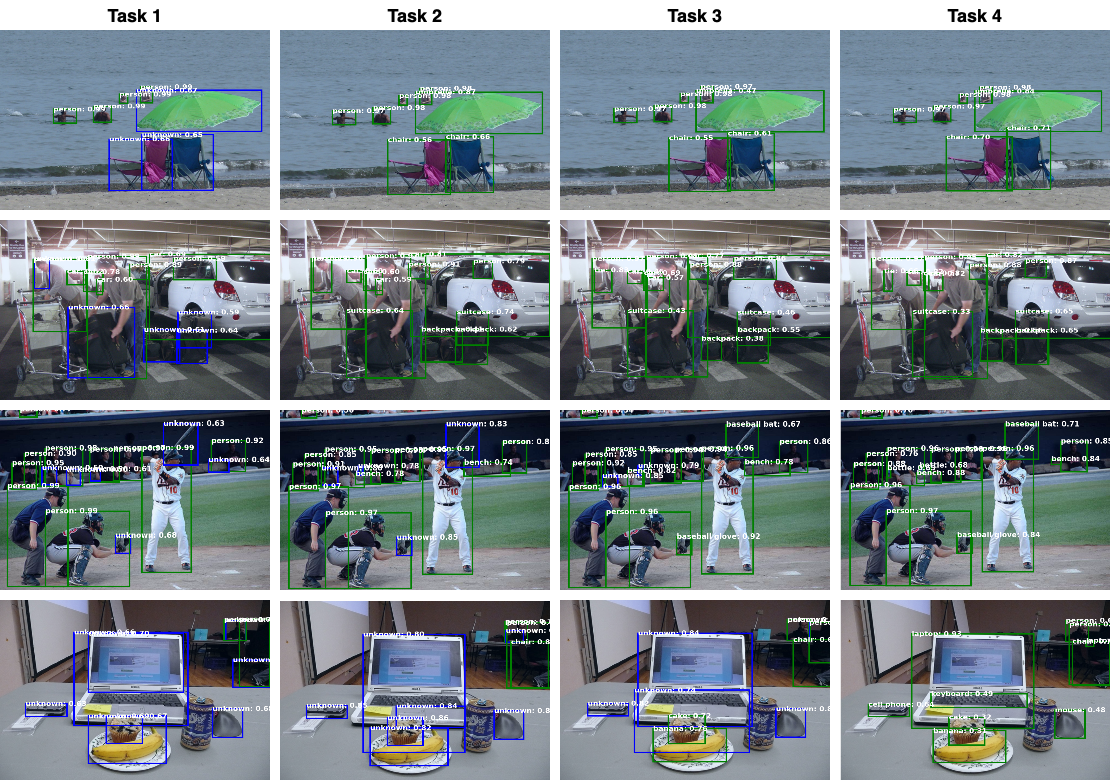}
  \caption{\textbf{Incremental Learning Performance Across Tasks.} Visualizations of predictions matched to ground truth objects across the tasks of S-OWOD. }
  \label{fig:inc_s}
\end{figure}

\begin{figure}
  \centering
  \includegraphics[width=0.99\linewidth]{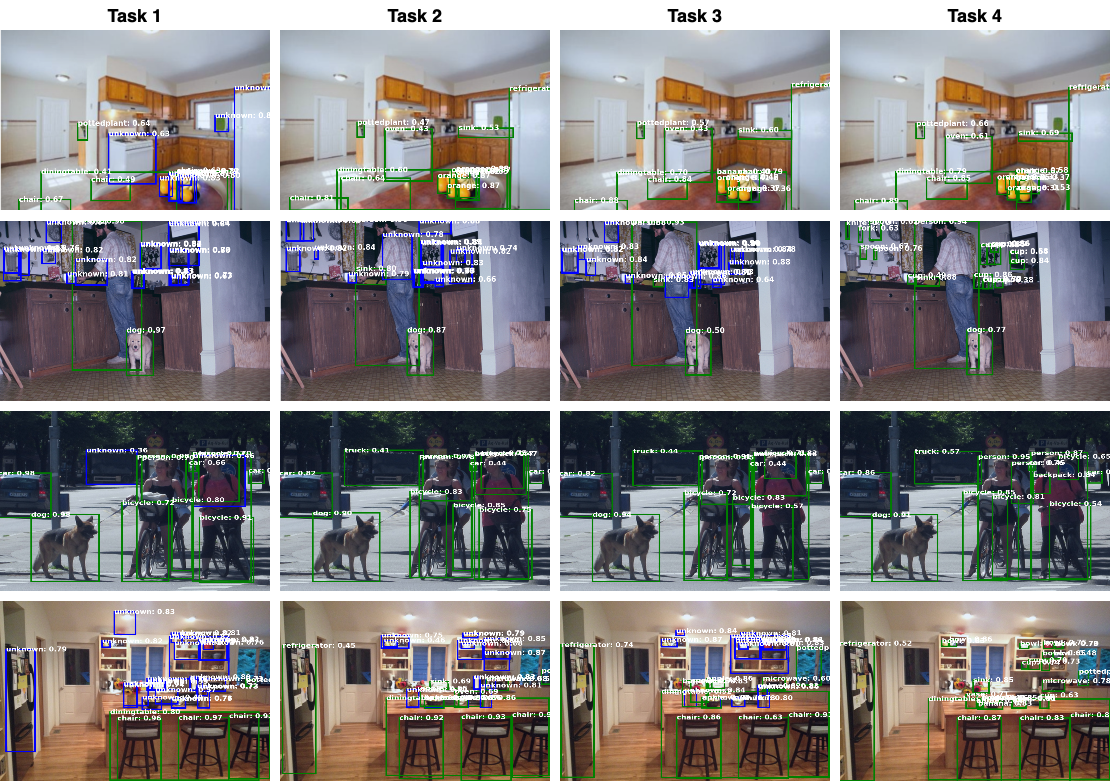}
  \caption{\textbf{Incremental Learning Performance Across Tasks.} Visualizations of predictions matched to ground truth objects across the tasks of M-OWOD. }
  \label{fig:inc_m}
\end{figure}

\boldparagraph{Incremental Learning} In Fig. \ref{fig:inc_s} and \ref{fig:inc_m}, we visualize the predictions of O1O across different tasks of S-OWOD and M-OWOD respectively. Different than previous top-k visualizations, we show the predictions matched to ground truth objects to showcase the development clearly. Although some classes are unknown in the first tasks, our model can still locate them. As the space of known categories expands, O1O learns to label them correctly without forgetting the previously learned classes, showing the effectiveness of our exemplar replay fine-tuning.

\end{document}